\documentclass[sigconf, 10pt]{acmart}

\usepackage[english]{babel}
\usepackage{blindtext}
\usepackage{caption}
\usepackage{subcaption}
\usepackage{color}
\usepackage{tabularx}
\usepackage{multirow}
\usepackage{multicol}
\usepackage{graphicx}
\usepackage{tabularray}
\usepackage{float}
\usepackage{xcolor} 
\usepackage{todonotes}
\usepackage{array} 
\usepackage{makecell} 
\usepackage{booktabs} 
\usepackage{geometry}
\usepackage{microtype}
\usepackage{booktabs} 
\usepackage{algorithm}
\usepackage{algorithmic}
\usepackage{enumitem}

\renewcommand\footnotetextcopyrightpermission[1]{} 
\setcopyright{none}

\newcommand{\name}{MobileExplorer\xspace}

\acmDOI{}

\acmISBN{}

\begin{document}

\title{MobileExplorer: Accelerating On-Device Inference for Mobile GUI Agents via Online Exploration}





\author{Runxi Huang}
\affiliation{\institution{Hong Kong University of Science and Technology}}
\email{rhuangbj@connect.ust.hk}

\author{Liyu Zhang}
\affiliation{\institution{Hong Kong University of Science and Technology}}
\email{lzhangcx@connect.ust.hk}

\author{Shengzhong Liu}
\affiliation{\institution{Shanghai Jiao Tong University}}
\email{shengzhong@sjtu.edu.cn}

\author{Xiaomin Ouyang}
\authornote{Corresponding author}
\affiliation{\institution{Hong Kong University of Science and Technology}}
\email{xmouyang@cse.ust.hk}


\begin{abstract}
Mobile graphical user interface (GUI) agents enable AI models to autonomously operate smartphones on behalf of users. However, most existing systems focus primarily on optimizing task accuracy and rely on cloud-hosted models for inference, which introduces privacy concerns and network-dependent latency. As a result, fully on-device deployment of mobile GUI agents remains underexplored.
We propose MobileExplorer, a new framework that accelerates on-device inference for vision-based mobile GUI agents via online exploration. The key idea is to exploit the long per-step reasoning time of vision-language models (VLMs) by performing lightweight, parallel exploration of UI elements. During model inference, the agent proactively probes semantically relevant UI elements and records exploration of these traces as structured memory. To ensure reliable execution in live mobile environments, we design a two-level rollback mechanism that robustly restores the initial UI state when fast but naive backtracking strategy fails. The collected exploration traces are then summarized into concise contextual hints and injected into the prompt for enhancing the subsequent reasoning step.
We evaluate MobileExplorer on multiple off-the-shelf devices using the AndroidWorld benchmark, as well as newly designed, more complex tasks and dynamic on‑device environments. MobileExplorer reduces the average number of reasoning steps and end-to-end latency by 23\% , while maintaining or improving task success rates by up to 5\%. A video demonstration of \name's performance in real world is available at \href{https://youtu.be/thK7MJmdlvM}{https://youtu.be/thK7MJmdlvM}.\footnote{The source code for \name will be released once the paper is accepted.}
\end{abstract}



\maketitle

\section{Introduction}

Mobile GUI agents have rapidly advanced with the progress of large language models (LLMs)~\cite{achiam2023gpt, dubey2024llama} and vision‑language models (VLMs)~\cite{bai2025qwen2, team2023gemini}, enabling end‑to‑end mobile task automation where understanding and planning are handled within a single model~\cite{wen2024autodroid}. 
These agents generally adopt two input modalities: \textit{text‑based}, which rely on accessibility trees~\cite{ding2024mobileagent, wen2024autodroid, wen2025autodroid, lee2024mobilegpt, dai2025advancing}, and \textit{vision‑based}, which operate directly on screenshots~\cite{wang2024mobile, ye2025mobile, wang2025ui, zhou2025mai, yan2025step}. Compared with text-based accessibility trees that mainly expose textual attributes, screenshots offer richer visual context—layout, spatial relations, and icons, enabling stronger visual grounding for complex interfaces~\cite{you2024ferret, li2022spotlight}. Consequently, vision‑based agents often outperform text‑based methods on challenging GUI tasks~\cite{li2022spotlight, hong2024cogagent}, and have become the dominant approach, which we focus on in this work.
However, most existing GUI agent systems--no matter text or vision-based input--run LLM/VLM reasoning models in the cloud, and only execute actions locally. This design requires uploading user interface data, creating significant privacy risks. Therefore, this highlights the growing need for fully on‑device mobile GUI agents that perform perception, reasoning, and action entirely locally.

Despite these advantages, building a fully on‑device vision‑based mobile GUI agent remains challenging. First, while VLMs provide stronger visual understanding than LLMs, they incur much higher computation and memory costs, making on‑device deployment difficult even for lightweight language models. For example, MAI‑UI‑2B~\cite{zhou2025mai} still requires about 40 seconds of latency on Samsumg Galaxy S24. Second, mobile GUI tasks demand fine‑grained visual grounding over complex interface elements (icons, layouts, text), which cannot be easily compressed or structured like accessibility‑tree inputs~\cite{wen2024autodroid}. Finally, mobile interfaces are highly dynamic due to pop‑ups, content changes, and device‑specific variations, often requiring multiple rounds of VLM reasoning. 

However, existing approaches for accelerating VLM inference in mobile GUI agents still face notable limitations. Multi‑step planning or script‑style execution~\cite{wen2024autodroid, wen2025autodroid} reduces the number of model calls by generating action sequences in advance, but such plans are brittle and often fail under dynamic UI changes. Verifier‑based pipelines~\cite{dai2025advancing} shift action generation to lightweight validation over candidate actions, yet their effectiveness depends heavily on candidate quality and still incurs nontrivial reasoning overhead. Token‑ or context‑pruning techniques~\cite{lin2025showui} reduces complexity of input data, but aggressive pruning may discard fine‑grained visual information which is essential for accurate GUI grounding. Moreover, most existing systems process reasoning and GUI interaction sequentially, leaving the long VLM inference time underutilized.

In this paper, we propose \name, a new on‑device mobile GUI agent framework that exploits the long VLM reasoning step latency to perform parallel online exploration. 
We observe that on-device mobile GUI agents exhibit a clear latency imbalance: UI perception and interaction are relatively lightweight, while on-device VLM reasoning incurs substantial latency.
Instead of remaining idle during model inference, the system proactively interacts with the current screen to gather task‑relevant information that strengthens subsequent reasoning steps, thereby reducing overall latency.
To enable efficient exploration within the model's reasoning latency, \name adopts a task relevance‑driven exploration strategy that prioritizes semantically important and clickable UI elements using lightweight text embeddings~\cite{reimers2019sentence}, with each selected element associated with precise coordinates. To ensure stable navigation, we design a robust rollback mechanism that reliably restores the interface to its initial UI state after each exploration attempt, preventing UI drift and guaranteeing that reasoning decisions are executed on the same screen state used for model inference.
The explored UI elements and interaction traces are then converted into structured, compact textual hints through lightweight template‑based summarization, where semantic labels derived from UI attributes are ranked and refined into concise prompts for the model. By externalizing and reusing exploration knowledge as a dedicated reasoning context, \name enhances per‑step reasoning accuracy to reduce both reasoning steps and end‑to‑end system latency.

We evaluate MobileExplorer on multiple off-the-shelf smartphones using the AndroidWorld benchmark~\cite{rawles2024androidworld}, along with newly designed complex tasks and dynamic on‑device environments. Under fully on-device execution, MobileExplorer preserves task success rates while reducing the average number of interaction steps and end-to-end latency by approximately 23\%.

The main contributions of this work are:
\begin{itemize}[leftmargin=12pt]
    \item We conduct an in-depth analysis of end-to-end execution latency in on-device mobile GUI agents, and identify that the long VLM reasoning time is underutilized in existing sequential interaction pipelines.
    \item We propose \name, a new on‑device mobile GUI agent framework that exploits model reasoning time to perform lightweight, parallel online exploration of UI elements, allowing the system to gather task‑relevant information to enhance subsequent reasoning steps.
    \item To enable effective exploration during model reasoning, we design a task relevance-driven exploration strategy that probes semantically meaningful and diverse UI elements, as well as a two‑level rollback mechanism that restores the initial UI state. The resulting exploration traces are then converted into structured prompt hints to enhance model reasoning.
    \item We evaluate \name on multiple commercial devices using the AndroidWorld benchmark and newly designed real‑world tasks. The results show that \name improves task success rates by up to 5\% while reducing both reasoning steps and end-to-end latency by 23\%.
    
\end{itemize}

\section{Related Work}
\label{sec:related work}


\noindent\textbf{Mobile GUI agent systems.}
Mobile GUI agents have been widely studied with both text-based and vision-based inputs. Early systems rely on LLMs operating on structured textual representations such as accessibility trees and action histories. For example, AutoDroid~\cite{wen2024autodroid} and AutoDroid-V2~\cite{wen2025autodroid} leverage planning and memory mechanisms for long-horizon GUI interactions, while V-Droid~\cite{dai2025advancing} adopts a verifier-based paradigm that selects actions from candidate UI elements. However, text-only representations miss rich visual cues such as icons, layout structures, and spatial relationships. Recent work therefore explores vision-based agents that operate directly on screenshots, including CogAgent~\cite{hong2024cogagent}, Mobile-Agent-V3~\cite{ye2025mobile}, MAI-UI~\cite{zhou2025mai}, and STEP-UI~\cite{yan2025step}. These systems improve GUI grounding through visual-region alignment and interaction trajectory training.
However, most existing agents rely on cloud-based inference and interact with devices remotely via ADB~\cite{adb_tool}, which introduces privacy risks and network dependency.

\noindent\textbf{On-device deployment of mobile GUI agents.}
A few studies explore running language-model-based agents directly on mobile devices~\cite{wen2024autodroid, wen2025autodroid}.
For example, AutoDroid~\cite{wen2024autodroid} performs local reasoning by representing GUI states as structured text and constructing UI transition graphs, while AutoDroid-V2~\cite{wen2025autodroid} reduces latency by generating action plans from app documentation. However, these methods rely on textual input and planning priors, which cannot be directly applied to vision-based agents that must reason over raw screenshots. Moreover, vision-based agents require VLM inference over high-dimensional visual inputs, introducing substantially higher computational overhead for on-device deployment. In contrast, we focuses on accelerating on-device inference for vision-based GUI agents by utilizing the reasoning time to perform lightweight exploration in parallel.

\noindent\textbf{Offline knowledge base construction for mobile GUI agents.}
Some mobile GUI agents improve decision making by incorporating prior knowledge from explored interfaces. For example, AutoDroid~\cite{wen2024autodroid} constructs a UI transition graph as structured memory, while AutoDroid-V2~\cite{wen2025autodroid} leverages app documentation and generated task samples for planning. Other works build such knowledge through offline exploration. GUI-explorer~\cite{xie2025gui} mines transition-aware knowledge from state--action traces, and LLM-Explorer~\cite{zhao2025llm} constructs reusable repositories of UI states and interaction graphs via large-scale app exploration. However, these approaches depend on offline-collected trajectories or pre-built knowledge bases, which are costly to construct and difficult to generalize to dynamic interfaces. In contrast, our work performs lightweight online exploration during inference without requiring offline knowledge construction, enabling adaptation to dynamic real-world mobile applications.


\section{A Motivation Study}
\label{sec:motivate_study}

\subsection{Background}

\subsubsection{Workflow of Mobile GUI Agents}

\begin{figure}
    \centering
    \setlength{\abovecaptionskip}{0.cm}
    \setlength{\belowcaptionskip}{0.cm}
    \includegraphics[width=1.0\linewidth]{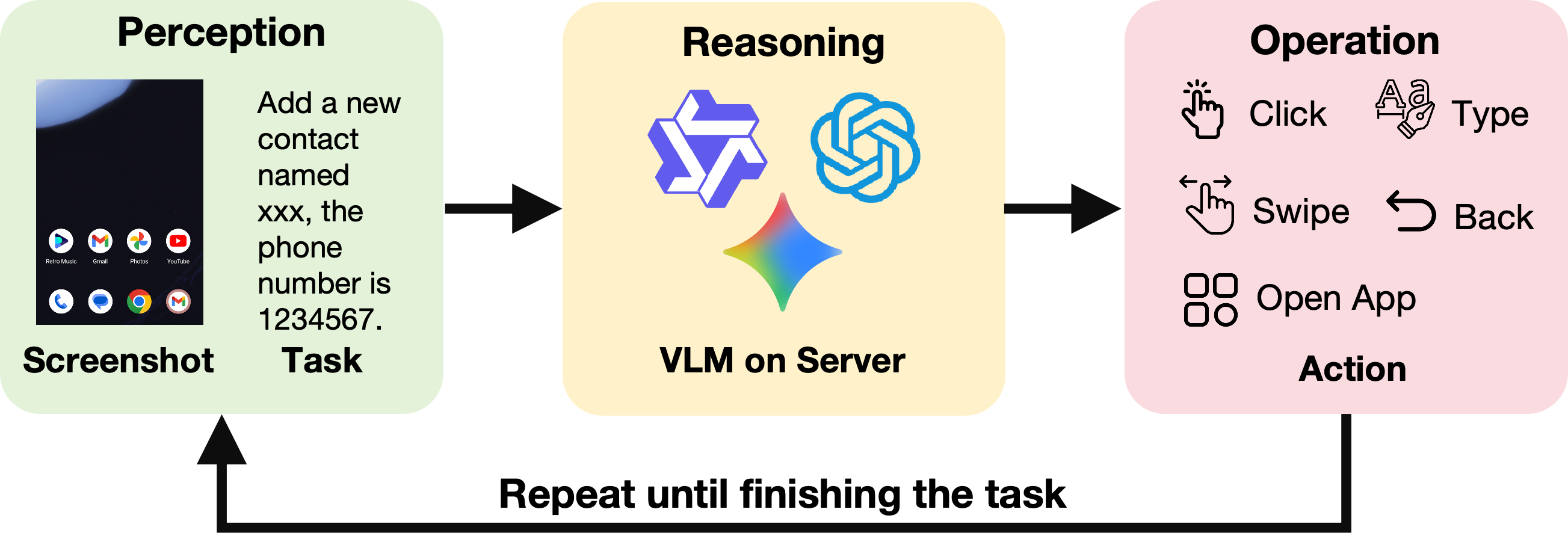}
    \caption{End-to-end workflow of a mobile GUI agent. The agent takes a screenshot as visual input, performs multimodal reasoning with VLMs, and outputs text that are mapped to executable GUI actions.}
    \label{fig:end_to_end}
    \vspace{-1em}
\end{figure}

Figure~\ref{fig:end_to_end} illustrates the typical end-to-end workflow of a mobile GUI agent, which consists of perception, reasoning, and operation. 
(1) Perception. Given the current screen, the agent first captures a screenshot that reflects the visual layout, icons, text, and spatial relationships among UI elements. (2) Reasoning and planning. Reasoning and planning. An image encoder converts the screenshot into visual embeddings. These embeddings are then combined with textual token--such as user instructions, task descriptions, or dialogue context—--nd fed into a VLM for reasoning. The VLM outputs executable GUI actions, such as tapping icons, typing text, scrolling, or navigating across apps. (3) Operation. The system executes these actions on the device and repeats the perception–reasoning–operation loop until the task is completed. For example, to accomplish the task “Turn on Wi‑Fi,” the agent must open the Settings app, scroll to locate the Wi‑Fi menu, and tap it.

\subsubsection{Vision-based Mobile GUI Agents}

\begin{figure}
    \raggedright   
    \setlength{\abovecaptionskip}{0.cm}
    \setlength{\belowcaptionskip}{-0.cm}
    \begin{subfigure}{0.49\linewidth}
            \setlength{\abovecaptionskip}{0.cm}
        \setlength{\belowcaptionskip}{0.cm}
        \raggedright   
        \includegraphics[width=1\linewidth]{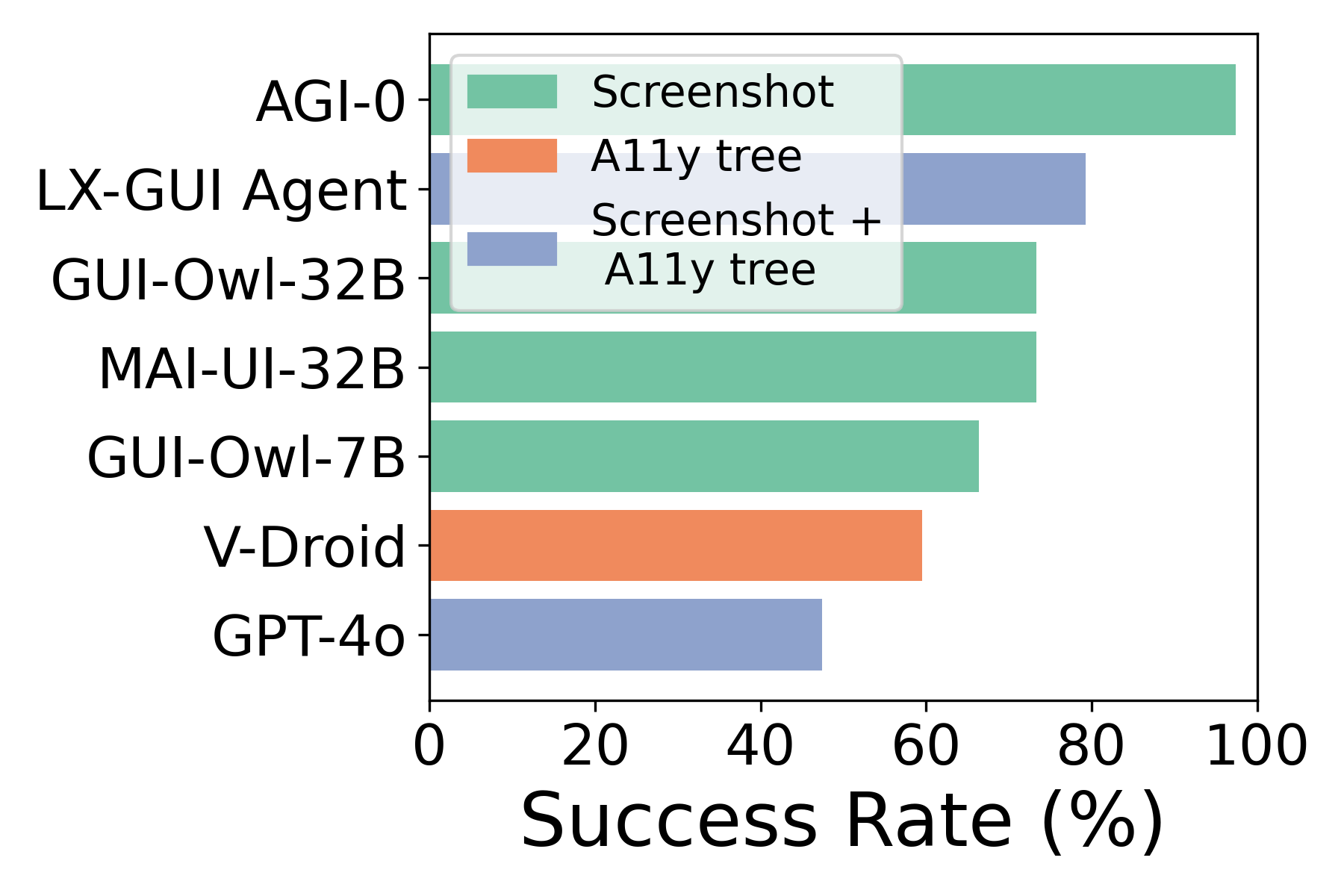}
        \caption{Top models on AndroidWorld~\cite{zhou2025mai, dai2025advancing, ye2025mobile, wang2025ui}.}
        \label{fig:motivation_benchmark}
    \end{subfigure}%
    \hspace{0.5mm}
    \begin{subfigure}{0.49\linewidth}
            \setlength{\abovecaptionskip}{0.cm}
        \setlength{\belowcaptionskip}{0.cm}
        \raggedright   
        \includegraphics[width=1\linewidth]{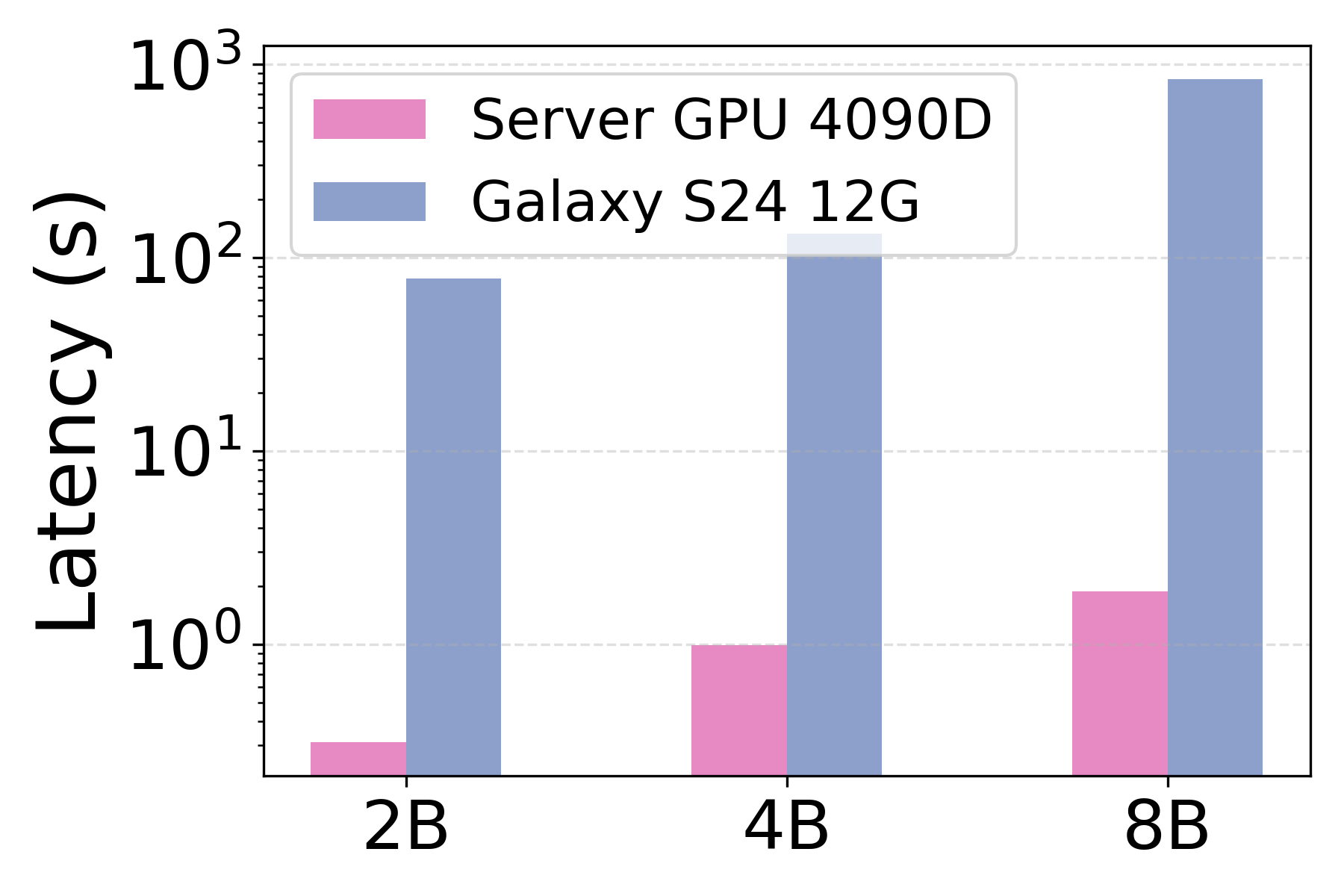}
        \caption{On-device VLM inference latency compared with server.}
        \label{fig:latency_comparison}
    \end{subfigure}%
    \caption{Vision-based agents dominate mobile GUI benchmarks but introduce substantial on-device inference latency.}
    \label{fig: motivation}
    \vspace{-1em}
\end{figure}

Compared with accessibility trees that mainly expose textual attributes, screenshots provide richer visual context such as layout structures, spatial relations, and icons, which are essential for understanding complex mobile interfaces~\cite{you2024ferret, li2022spotlight}. As shown in Fig.~\ref{fig:motivation_benchmark}, the top-performing agents on the AndroidWorld benchmark are predominantly vision-based. For example, strong agents such as GUI-Owl~\cite{ye2025mobile} and MAI-UI~\cite{zhou2025mai} rely on screenshot inputs for perception. In contrast, approaches that depend mainly on accessibility trees rarely appear among the leading entries. This difference is particularly evident in tasks that require recognizing icons or spatial layout. These observations indicate that vision is a fundamental capability for reliable mobile GUI agents.

However, visual perception also introduces significant system challenges. Vision–language models incur substantially higher computational cost than LLMs, leading to large latency when deployed directly on mobile devices. To quantify this overhead, we measure the inference latency of representative VLMs on a Samsung Galaxy S24 smartphone with 12GB RAM. All models are executed using \texttt{llama.cpp} and quantized to Q8 to enable on-device execution. Each inference processes a screenshot with a resolution of $540\times1200$. The results in Fig.~\ref{fig:latency_comparison} show that VLM inference introduces substantial latency on mobile hardware, and the delay increases rapidly with model size. These observations highlight a key systems challenge: although visual perception is essential for mobile GUI agents, directly executing VLM reasoning on-device can lead to significant end-to-end latency.

\subsection{Understanding Latency of On-device Mobile GUI Agent Systems}

\begin{figure}
    \raggedright   
    \begin{subfigure}{0.49\linewidth}
        \centering
        \includegraphics[width=1\linewidth]{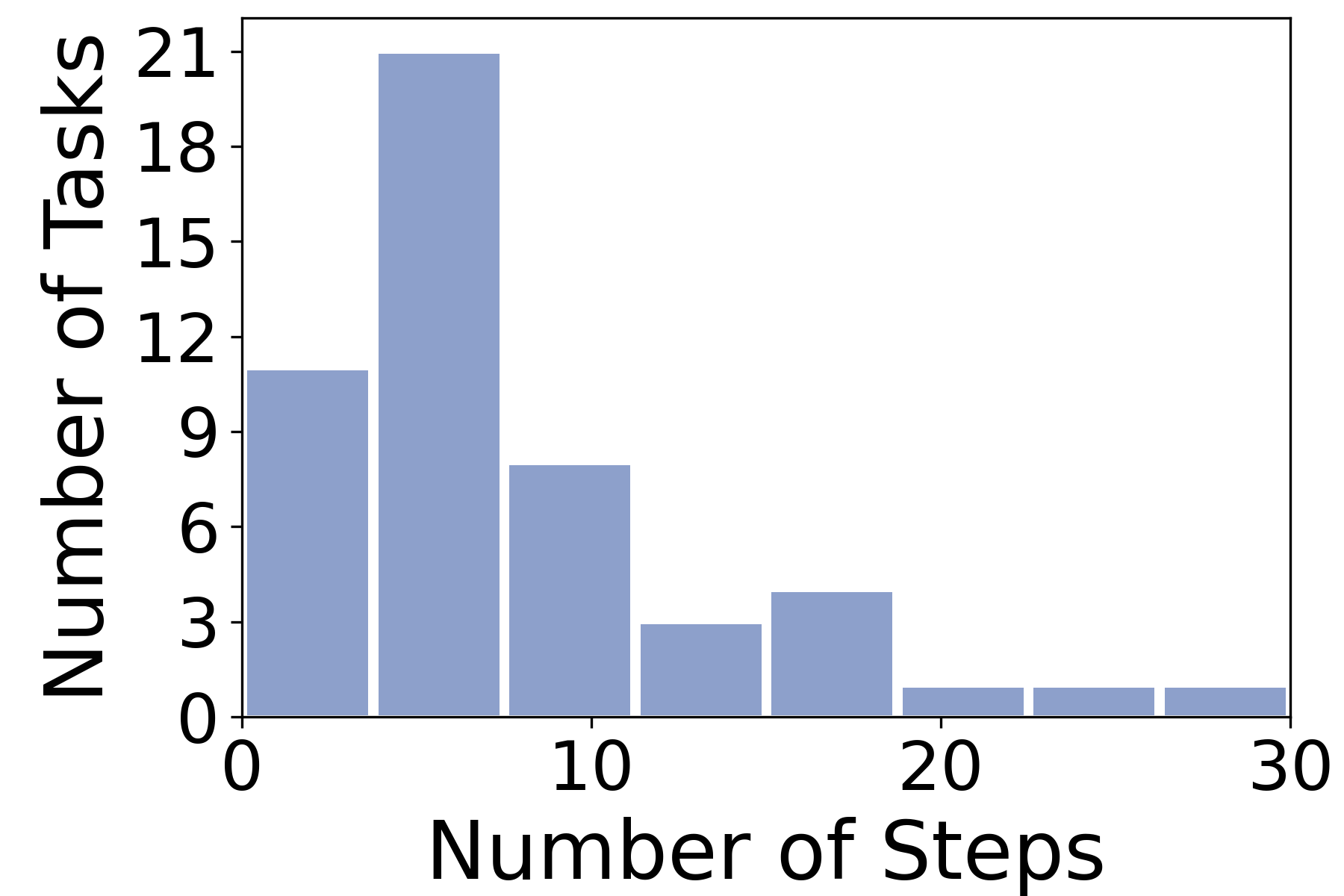}
        \caption{Number of steps of tasks in AndroidWorld~\cite{rawles2024androidworld}.}
        \label{fig:android_steps}
    \end{subfigure}%
    \hspace{0.01\linewidth}
    \begin{subfigure}{0.49\linewidth}
        \raggedright   
        \includegraphics[width=1\linewidth]{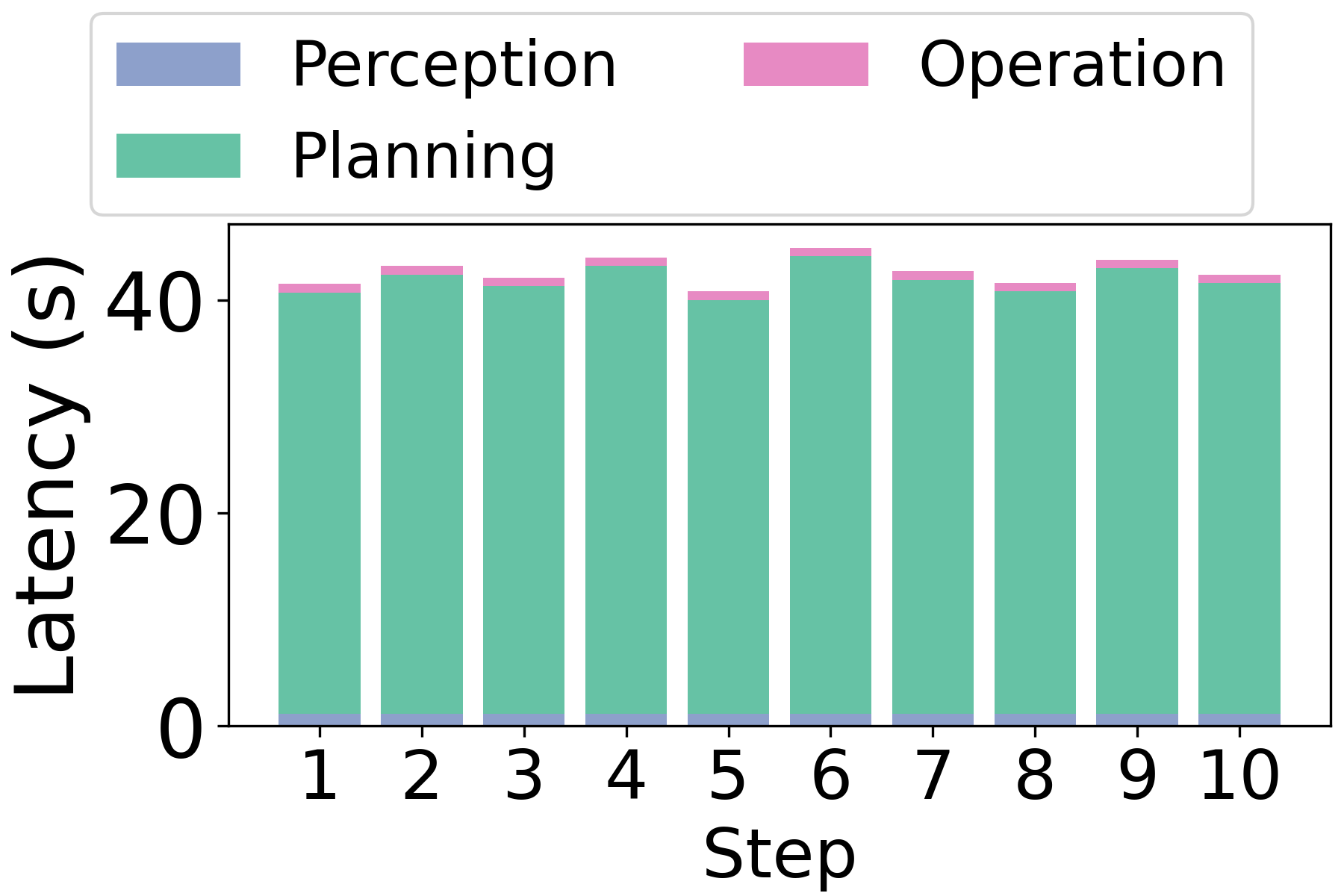}
        \caption{Latency across different steps of a task.}
        \label{fig:step_latency}
    \end{subfigure}%
    \caption{Latency of Mobile GUI Agent Systems. There are lots of idle reasoning time of model inference across different steps.} 
    \label{fig:overhead evaluation}
    \vspace{-1em}
\end{figure}

\begin{figure*}
    \centering
    \setlength{\abovecaptionskip}{0.cm}
    \setlength{\belowcaptionskip}{-0.cm}
    \includegraphics[width=\textwidth]{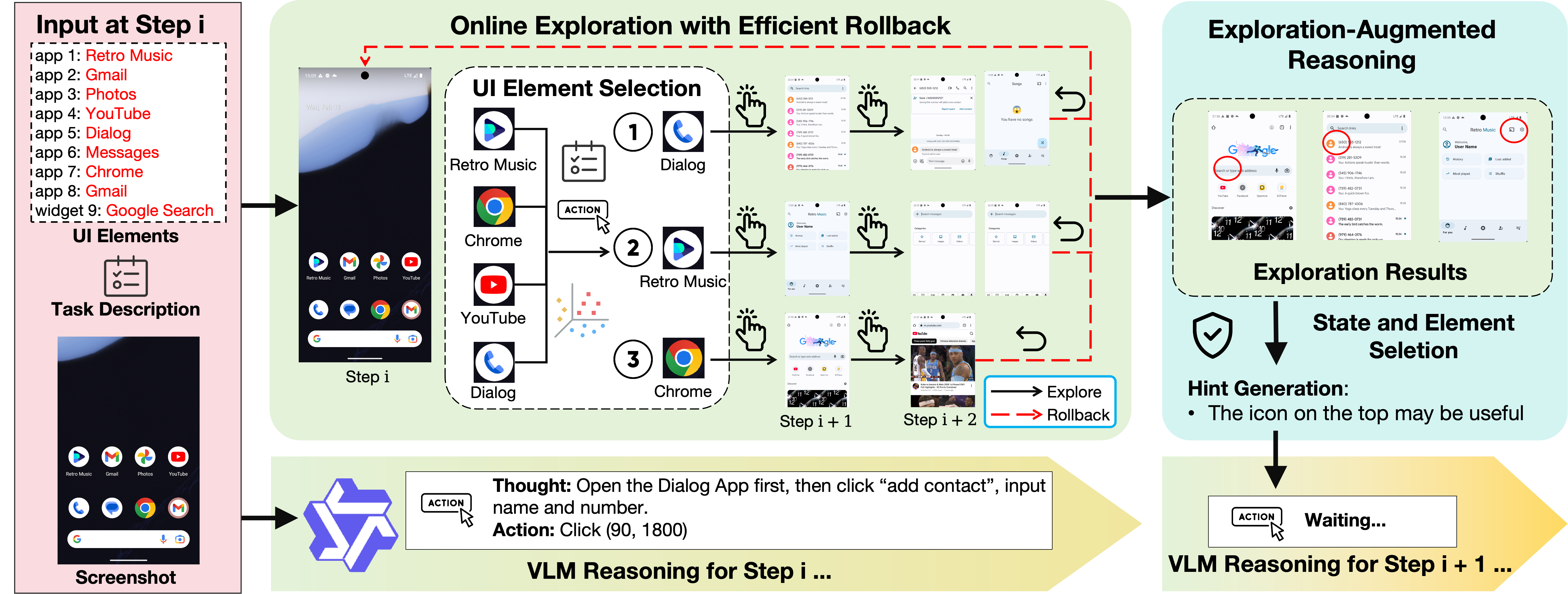}
    \caption{System overview of \name. During each reasoning step, the system performs lightweight online exploration in parallel with VLM inference, uses a robust rollback mechanism to restore the initial UI state, summarizes the discovered screen information into compact hints, and feeds them into the next reasoning step. }
    \label{fig:system_overview}
\end{figure*}



Visual reasoning introduces a critical systems bottleneck on mobile devices due to the high computational cost of VLM inference on resource-constrained hardware. To quantify this overhead, we conduct a preliminary measurement using a task from the AndroidWorld benchmark—\textit{recording and saving an audio file on a commercial smartphone}. The experiment is performed on a Samsung Galaxy S24 (12GB RAM) by running a 2B VLM (MAI-UI~\cite{zhou2025mai}) fully on-device. Fig.~\ref{fig:android_steps} shows the latency breakdown of a single decision step. Even for this simple task, each step incurs substantial planning latency (tens of seconds), while perception and UI operations account for only a small fraction of the total time. This indicates that most of the system latency is spent waiting for the VLM to complete reasoning before issuing the next action. Fig.~\ref{fig:step_latency} further measures the inference latency of VLMs with different model sizes on the same device.

Moreover, mobile GUI tasks are inherently multi-step and sequential. Fig.~\ref{fig:step_latency} shows that completing a task often requires many interaction steps, with some tasks exceeding 15–20 actions. Consequently, the large per-step reasoning latency accumulates across the interaction trajectory, leading to extremely high task completion time on resource-constrained devices. These observations suggest that the inefficiency of on-device GUI agents stems not from a single slow inference, but from the combination of long-horizon decision making and heavy per-step reasoning cost.

\subsubsection{Opportunity for online exploration.}
The above preliminary analysis also reveal a key opportunity for accelerating on-device mobile GUI agent systems. Each VLM reasoning step takes tens of seconds, whereas lightweight UI interactions—such as taps or screen transitions—typically require only 1–2 seconds. Yet existing mobile GUI agents leverage a sequential pipeline: the system waits for the model's next action before interacting with the UI, leaving the device idle during long inference periods. 
We argue that this idle interval can instead be exploited. Rather than treating reasoning latency as unavoidable overhead, the system can perform lightweight online exploration in parallel with model inference. By probing selected UI elements and observing the resulting screen transitions, the agent can uncover additional UI context that reduces uncertainty in later decisions and shortens the overall interaction trajectory.

\subsection{Summary}

We now summarize the key findings from our motivation study.

\begin{itemize}[leftmargin=12pt]
    \item  Although vision‑based mobile GUI agents dominate realistic mobile benchmarks, they incur substantial end‑to‑end latency for on-device deployment due to long VLM inference time and the accumulation of delays across multi‑step reasoning.
    
    \item Since UI interactions are much faster than model reasoning, the large idle intervals that arise during model inference provide opportunities for lightweight parallel exploration.
\end{itemize}

\section{System Overview}
\label{sec:system_overview}

We propose \name, a fully on-device vision-based mobile GUI agent that overlaps VLM reasoning with lightweight online exploration, as illustrated in Fig.~\ref{fig:system_overview}. Unlike conventional GUI agents that follow a sequential perception–reasoning–operation pipeline, \name treats the long VLM reasoning phase as an exploitable time window and performs lightweight UI exploration concurrently on the same device to gather additional task-relevant UI context. All processes, including perception, reasoning, exploration, and interaction, are executed locally on the smartphone, preserving user privacy and eliminating any need for cloud communication.

At each step $i$, the agent captures the current screenshot together with the task description and feeds them to the VLM for action reasoning. Meanwhile, instead of waiting for the model output, the system parses UI elements from the current screen and computes semantic similarity between candidate interactive elements and the task description using a lightweight text embedding model. Based on this relevance score, the agent selects a small set of clickable and diverse UI elements and performs lightweight probing actions (e.g., tapping elements and observing screen transitions) to explore potential UI branches. To maintain consistency with the main interaction trajectory, the system records exploration traces and reliably restores the interface to the original UI state where exploration started before executing the next decision. The exploration outcomes, including visited screens and discovered UI semantics, are then summarized into compact textual hints through lightweight template-based summarization. These hints are appended to the prompt of the next reasoning step, enabling the VLM to incorporate newly discovered contextual information instead of relying solely on the original screenshot.

Overall, \name forms a partially parallel execution pipeline: VLM reasoning operates on the current step $i$, while exploration simultaneously collects contextual knowledge that can assist future reasoning steps. By overlapping model inference with lightweight UI exploration, the system utilizes otherwise idle reasoning time to gather additional UI evidence without increasing the number of model invocations. This design improves decision accuracy and reduces unnecessary trial-and-error interactions in long-horizon mobile GUI tasks, ultimately reducing the overall end-to-end latency while remaining fully deployable on commodity smartphones.

\section{Design of \name}

\subsection{Problem Formulation}

We consider an on-device vision-based mobile GUI agent that interacts with an application to finish a user-defined task through multi-step interactions (including perception, reasoning and operations). At step $i$, the agent observes the current UI screenshot $s_i$ and receives a task description $\mathcal{G}$ 
Due to the partial observability of mobile interfaces, the full UI structure cannot be known a priori and can only be revealed through interaction.

\emph{State and Action Space.}
At step $i$, the agent observes the current screenshot $s_i$ and parses the set of interactive UI elements:
\begin{equation}
    \mathcal{A}_i = \{a_i^{(1)}, a_i^{(2)}, \dots, a_i^{(K_i)}\}.
\end{equation}
Each action corresponds to interacting with a UI element (e.g., click or scroll), which may trigger a transition to a new UI state.

\emph{Latency-Constrained Interactions.}
On-device VLM inference incurs significant latency. Let $\tau_i^{\text{vlm}}$ denote the reasoning latency at step $i$. During this period, the device remains idle from the interaction perspective, providing an opportunity to perform auxiliary operations without increasing user-perceived delay.

\emph{System Objective.}
The objective of \name is to maximize the probability of completing a task while minimizing interaction cost under on-device latency constraints:
\begin{equation}
\max_{\pi} \Pr(v_N \in \mathcal{V}_{\text{goal}}(\mathcal{G})),
\quad \text{s.t.} \quad
\sum_i \tau_i^{\text{interact}} \le T,
\end{equation}
where $T$ is the latency budget and $\pi$ denotes the interaction policy.

To achieve this objective, \name leverages the otherwise idle reasoning latency to perform lightweight online exploration and gather additional UI context without increasing the overall execution delay. This design introduces several challenges, including selecting informative interactions under strict time constraints, summarizing discovered UI states into compact reasoning context, and reliably restoring the original UI state after exploration.

\subsection{Task Relevance-driven Exploration}

\begin{figure}
    \centering
    \setlength{\abovecaptionskip}{0.cm}
    \setlength{\belowcaptionskip}{0.cm}
    \includegraphics[width=1.0\linewidth]{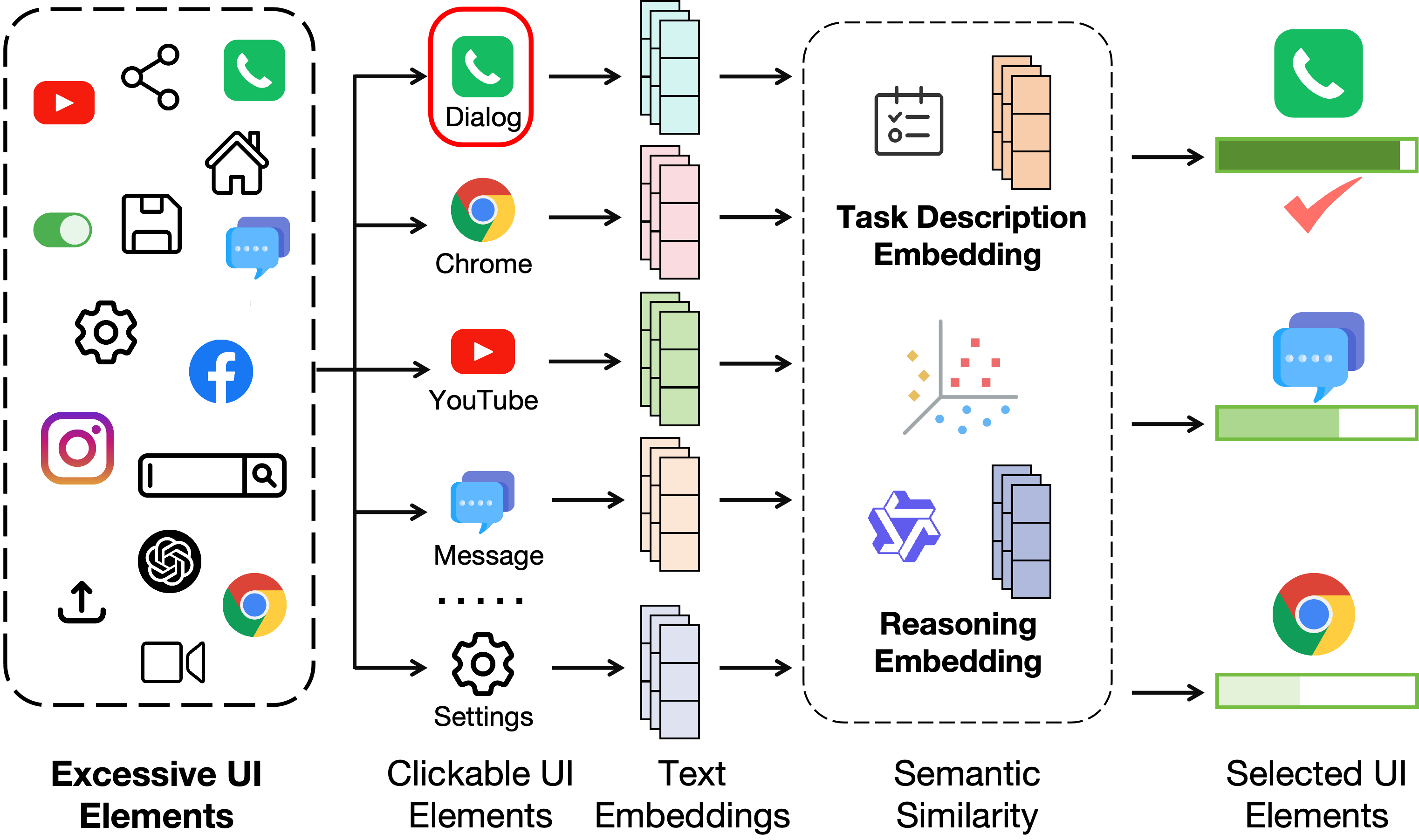}
    \caption{Task relevance-driven exploration. During each VLM reasoning step, the system ranks clickable UI elements by task relevance and performs short exploratory probes within model reasoning time.}
    \label{fig:exploration_strategy}
\end{figure}

As motivated by Section~\ref{sec:motivate_study}, the goal of online exploration is to proactively gather additional UI context during the VLM inference to improve subsequent reasoning. During model reasoning, the agent interacts with selected UI elements to uncover hidden UI branches and collect auxiliary information that may not be visible from the current screenshot alone. 


\subsubsection{Challenges.}
First, online exploration must operate strictly within the latency of the VLM inference at each reasoning step to avoid introducing additional delay. Consequently, exhaustive traversal of the UI graph is infeasible on resource-constrained mobile devices. In addition, a typical mobile screen may contain many clickable UI elements associated with different applications or actions, making it difficult to determine which interactions are most informative for the current task. Therefore, the exploration mechanism must both prioritize task-relevant elements and avoid repeatedly probing previously explored UI branches.

\subsubsection{Exploration Strategy.} 
To efficiently identify interaction candidates, the exploration module leverages the accessibility tree, which exposes structured UI elements with precise coordinates and interaction attributes while remaining far more lightweight than vision‑based UI parsing. At the beginning of step $i$, the exploration controller parses the accessibility tree of the current screen $s_i$ and extracts the clickable element set:
\begin{equation}
    \mathcal{A}_i = \{a_i^{(1)}, a_i^{(2)}, \dots, a_i^{(K_i)}\}.
\end{equation}

Each element $a_i^{(k)}$ is converted into a lightweight textual representation (e.g., text label, content description, resource identifier). 
As illustrated in Fig.~\ref{fig:exploration_strategy}, a typical mobile screen may contain multiple candidate UI elements corresponding to different applications or actions. 
For example, on a home screen the accessibility tree may expose elements associated with apps such as Phone, Chrome, or YouTube. 
Given a task description $\mathcal{G}$, the system computes semantic embeddings for both the task descriptions and each UI element, and ranks candidates based on their similarity. 
Elements whose semantics are more relevant to the task (e.g., launching a browser for a web search task) receive higher exploration priority, while unrelated elements receive lower scores.
Specifically, we compute the semantic similarity between the element representation and the task description $\mathcal{G}$ using a lightweight embedding model, producing a task relevance score

\begin{equation}
    r(a_i^{(k)}) = \text{sim}\!\left(e(a_i^{(k)}), e(\mathcal{G})\right),
    \label{eq:sim}
\end{equation}

where $e(\cdot)$ denotes a lightweight text embedding model used to encode both UI element descriptions and the task goal into semantic vectors, and $\text{sim}$ denotes the cosine similarity between embeddings.

To avoid repeatedly probing the same UI branches, the system incorporates exploration history. Let $\mathcal{H}_i$ denote the set of previously visited UI elements. The final exploration priority score is defined as

\begin{equation}
    S(a_i^{(k)}) = r(a_i^{(k)}) - \lambda \cdot \mathbf{1}[a_i^{(k)} \in \mathcal{H}_i],
\end{equation}

where $\lambda$ is a penalty weight that discourages revisiting explored elements.

The exploration controller then probes candidates in descending order of $S(\cdot)$. Each probe consists of executing a click action and observing the resulting screen transition. The exploration proceeds sequentially up to a bounded depth $d$, meaning that at most $d$ exploratory actions are executed before returning to the starting UI state. The resulting screen is parsed again to collect additional UI elements and contextual information. Because each probe is short and bounded, multiple probes can be executed within the same reasoning window.

Exploration terminates immediately once the time budget $\tau_i^{\text{explore}}$ is exhausted. The discovered UI elements, screens, and contextual observations are stored as exploration context $\mathcal{C}_i$ for subsequent reasoning steps. As a result, \name can efficiently gather task-relevant UI context within the VLM reasoning window while avoiding redundant exploration, enabling informative probing of UI branches without increasing end-to-end latency.


\subsection{Robust Rollback via Two-level Checking}

Online exploration in mobile GUI environments presents significant challenges compared with offline exploration in simulators. In offline settings, agents can freely duplicate or clone environment states before exploring alternative branches. In contrast, \name operates on a live mobile device, where every exploratory action directly alters the real UI state. Without reliable state recovery, exploration may drift the system away from the screen on which the reasoning decision was made, leading to inconsistent or incorrect execution. Therefore, developing a robust rollback strategy is essential for effective online exploration in  dynamic, real‑time settings. 

\subsubsection{Challenges}
Rollback in mobile GUI exploration is challenging due to two key properties of mobile interfaces. First, many UI transitions lack a true inverse operation. For example, a \textit{back} action may close a dialog, dismiss a keyboard, or skip intermediate screens rather than returning to the exact previous state, making simple reversal of exploration actions unreliable. Second, UI states can evolve between interactions even when following the same navigation path.  At step $i$, the agent only observes the rendered screenshot $s_i$ corresponding to the current UI state $s_i$, while the underlying application state and navigation stack remain hidden.

These properties make naïve backtracking insufficient. The \textit{back} operation may skip multiple pages, modal dialogs may insert temporary layers with non‑standard navigation behavior, and tapping certain elements may trigger permission prompts or confirmation windows not present before. Moreover, asynchronous updates—--uch as notification banners or dynamically loaded content--can alter the interface, producing different screenshots even along identical navigation paths. Therefore, reliably restoring the exact starting state requires a robust rollback mechanism that tolerates minor visual variations while ensuring the agent returns to the correct UI context.

\begin{figure}
    \centering
    \setlength{\abovecaptionskip}{0.cm}
    \setlength{\belowcaptionskip}{0.cm}
    \includegraphics[width=1.0\linewidth]{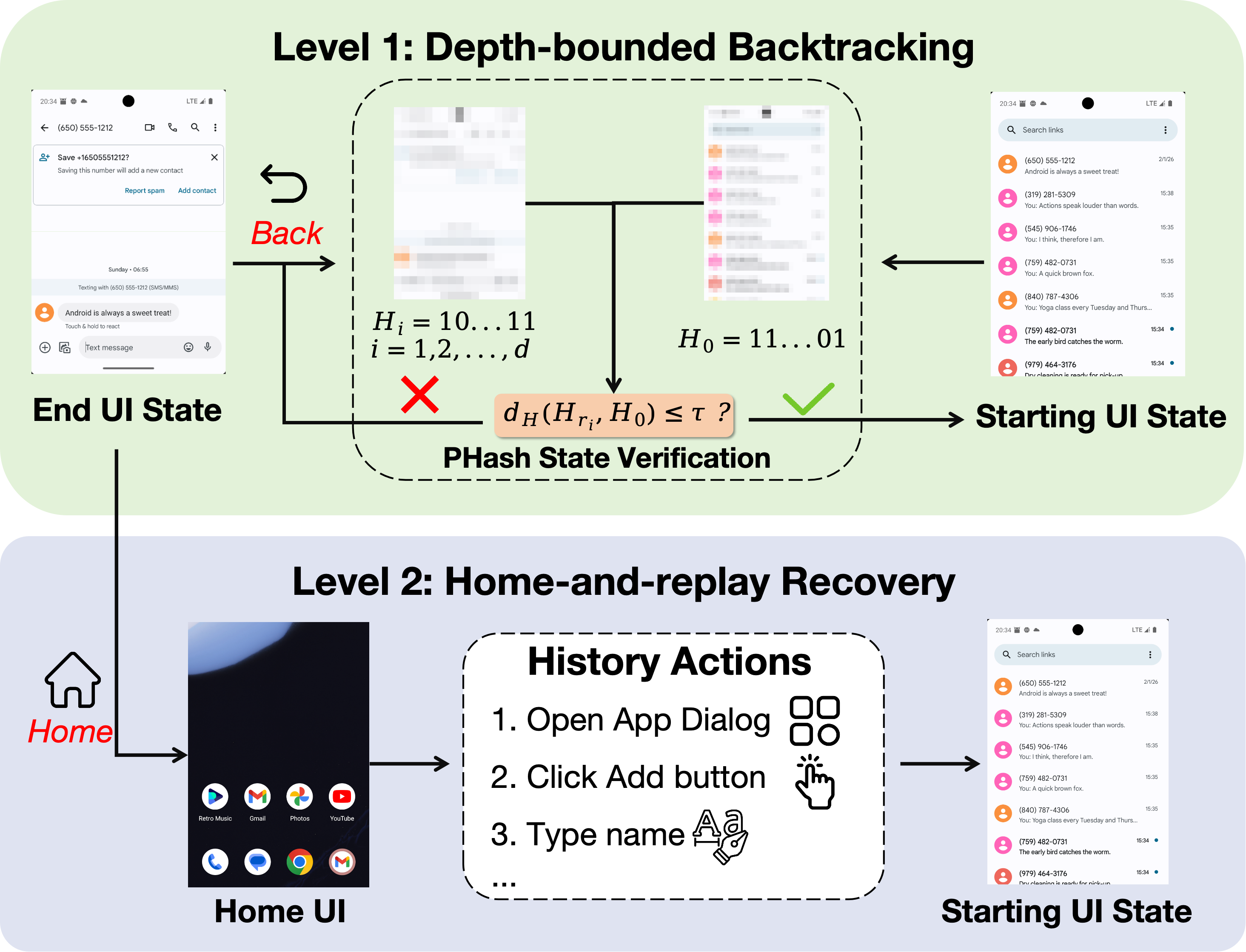}
    \caption{Two-level rollback strategy with perceptual-hash state verification and home-and-replay recovery.}
    \label{fig:exploration_return}
\end{figure}

\subsubsection{Two-Level Rollback Strategy}

To ensure reliable recovery under these conditions, \name adopts a two-level rollback strategy illustrated in Fig.~\ref{fig:exploration_return}. 
The two levels serve complementary roles. The first level performs fast depth-bounded backtracking, which efficiently restores the original UI state in most cases. However, due to irreversible UI transitions and dynamic interface changes, backtracking may occasionally fail to reach the exact starting screen. The second level therefore provides a deterministic recovery mechanism that reconstructs the UI state through action replay. Together, these two levels ensure both efficiency and robustness during exploration.

\emph{Level-1: Depth-Bounded Backtracking.}
Exploration is performed with a bounded depth $d$. Let the exploration start from UI state $s_i$. Meanwhile, the system records the interaction trace that led to this exploration state. We compute the perceptual hash of the starting screen as

\begin{equation}
H_0 = H(s_i),
\end{equation}

where $H(\cdot)$ denotes the pHash function. We use screenshot-based verification instead of accessibility-tree because capturing screenshots on mobile devices is significantly faster than retrieving the accessibility tree. Frequent screen verification is required during rollback, and screenshot capture introduces much lower latency. Perceptual hashing further allows efficient comparison between screens while tolerating minor visual differences caused by dynamic UI elements. After finishing an exploratory branch, the agent issues the \texttt{back} action up to $d$ times to return to the previous screens. After each rollback step $k$, the current screen $s_{r_k}$ is captured and verified against the starting screen:

\begin{equation}
d_H(H(s_{r_k}), H_0) \le \tau,
\end{equation}

where $d_H(\cdot)$ denotes the Hamming distance between perceptual hashes and $\tau$ is a small threshold allowing minor UI variations (e.g., scrolling offsets or dynamic content). If the condition holds, the system concludes that the original UI state has been successfully restored.

\emph{Level-2: Home-and-Replay Recovery.}
If the Level-1 rollback fails (e.g., due to irreversible UI transitions or navigation inconsistencies), the system performs a fallback recovery. For example, during exploration an interaction may trigger a confirmation dialog or permission prompt that was not previously visible. In such cases, issuing the \textit{back} action may dismiss the dialog and return to an earlier screen rather than the exact exploration starting state. As a result, the agent cannot reliably restore the original UI state through simple backtracking.

To recover from such situations, the system falls back to a deterministic reconstruction procedure. Let the interaction trace that led to the exploration starting state be

\begin{equation}
    \Pi_i = (a_1, a_2, \dots, a_i),
\end{equation}

where $a_k$ denotes the action executed at step $k$. The agent first returns to the Home screen and then deterministically replays the recorded trace $\Pi_i$ to reconstruct the UI state $s_i$. Since exploration depth is small in practice, the replay sequence is typically short and introduces negligible overhead compared to VLM inference latency.

\begin{algorithm}
\caption{Exploration with Rollback}
\label{alg:explore_rollback}
\small
\begin{algorithmic}[1]
\REQUIRE Task $\mathcal{G}$, start screen $s_i$, clickable set $\mathcal{A}_i$, depth $d$, budget $\tau_i$, threshold $\delta$
\ENSURE Exploration hints $\mathcal{C}_i$ and restored UI state $s_i$

\STATE $\mathcal{C}_i \leftarrow \emptyset$
\STATE $H_0 \leftarrow H(s_i)$ \hfill (compute pHash of start screen)

\STATE Rank candidates $\mathcal{A}_i^{\text{cand}}$ by semantic similarity to $\mathcal{G}$

\FORALL{$a \in \mathcal{A}_i^{\text{cand}}$ \textbf{within} $\tau_i$}
    \STATE Interact with $a$ and explore transitions up to depth $d$
    \STATE Record discovered screens / elements into $\mathcal{C}_i$
    \STATE Issue \texttt{back} for $d$ steps \hfill (Level-1 rollback)
    \STATE Capture current screen $s'$
    \STATE $H' \leftarrow H(s')$
    \IF{$d_H(H', H_0) > \delta$}
        \STATE Go \texttt{home} and replay reasoning trace $\Pi_i$ \hfill (Level-2 recovery)
    \ENDIF
\ENDFOR
\STATE \textbf{return} $\mathcal{C}_i$
\end{algorithmic}
\end{algorithm}

Algorithm~\ref{alg:explore_rollback} outlines the overall exploration process with rollback. This two-level design provides a fast rollback path in the common case (Level-1) while guaranteeing correctness under non-deterministic UI behaviors (Level-2). As a result, exploration can safely probe alternative UI branches without permanently altering the agent's main decision trajectory.

\subsection{Exploration-Augmented Reasoning}

Online exploration allows the agent to proactively gather additional UI context during the VLM inference window. However, exploration is performed before the reasoning decision is finalized. As a result, the reasoning action may lead the agent to a different screen than those visited during exploration. Blindly injecting all exploration results into the prompt may therefore introduce irrelevant information and increase reasoning latency. 

\subsubsection{Challenges.} Two key challenges arise in this process. First, exploration observations may become misaligned with the UI state on which reasoning is performed. Exploration starts from screen $s_i$, but the executed reasoning action may lead to a new screen $s_{i+1}$, making some exploration trajectories no longer relevant to the current decision context. Second, exploration may discover many UI elements within a short time window. Directly injecting all discovered elements into the prompt may introduce noise and unnecessarily increase prompt length, negatively affecting both reasoning accuracy and latency.

\subsubsection{UI State Selection.} During exploration, the agent may visit multiple intermediate UI states. To ensure that exploration results correspond to the current reasoning context, \name records lightweight exploration observations and verifies their relevance to the current screen.

\emph{Exploration Observations.}
Each visited screen is recorded as a compact observation consisting of the perceptual hash (pHash) of the screenshot and the parsed UI elements extracted from the accessibility tree. These observations capture screen identity and UI structure while avoiding large memory overhead.

\emph{Step Alignment.}
After the reasoning action is executed and the agent reaches screen $s_{i+1}$, the system computes the pHash of the current screen and retrieves exploration observations with similar visual structure. Formally, the matched observation set is defined as

\begin{equation}
\mathcal{O}_i^{\text{match}} =
\{o_j \in \mathcal{O}_i \mid d(h_{i+1}, h_j) < \delta \},
\end{equation}

where $d(\cdot)$ denotes the Hamming distance between perceptual hashes. This step filters out exploration results that correspond to unrelated UI states.

\begin{figure}
    \centering
    \setlength{\abovecaptionskip}{0.cm}
    \setlength{\belowcaptionskip}{0.cm}
    \includegraphics[width=1.0\linewidth]{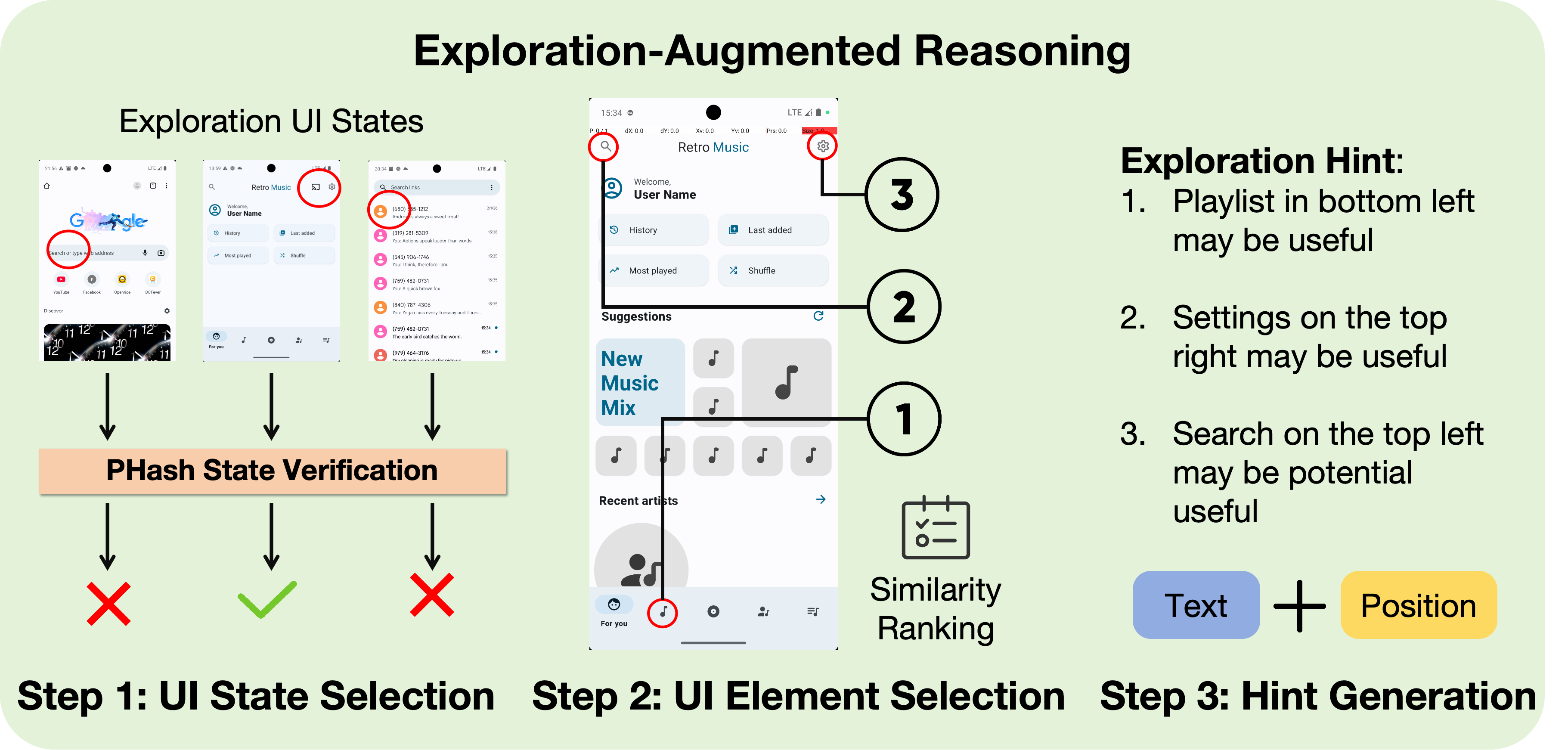}
    \caption{Exploration-augmented reasoning. The agent explores nearby UI states during inference, ranks discovered elements by task relevance, and generates textual hints to guide the next reasoning step.}
    \label{fig:exploration_augment}
\end{figure}

\subsubsection{UI Element Selection.} 
The matched exploration states may still contain multiple interactive UI elements corresponding to different applications or actions. Directly injecting all discovered elements into the prompt may introduce irrelevant information and mislead the model's reasoning. 

Therefore, \name selects only a small set of task-relevant elements from the explored screens. Specifically, for each element $a_i^{(k)}$, the system computes a task relevance score using the semantic similarity defined in Eq.~\ref{eq:sim}, which measures the similarity between the element representation and the task goal $\mathcal{G}$. Elements with higher similarity scores are considered more relevant to the task and are prioritized for hint construction, while unrelated elements are filtered out.

\subsubsection{Hint Generation.}
From the selected elements, \name summarizes exploration results into concise textual hints that describe potentially useful UI locations. These hints form the exploration context $\mathcal{C}_i$, which is appended to the prompt for the next reasoning step.

The VLM therefore reasons over an augmented input consisting of the current screenshot, the task instruction, and the exploration context. This design allows the agent to leverage additional UI knowledge discovered during exploration while keeping the reasoning input compact and relevant for on-device deployment.
\section{Evaluation}

\subsection{Experiment Setup}

\noindent\textbf{Benchmark.}
We evaluate \name on AndroidWorld~\cite{rawles2024androidworld}, a task-oriented mobile GUI interaction benchmark that requires multi-step reasoning over real smartphone interfaces (details in Table~\ref{tab:androidworld}). Unlike static GUI datasets such as Android-in-the-Wild~\cite{rawles2307android} and AndroidControl~\cite{li2024effects}, which rely on offline interaction traces, AndroidWorld provides an interactive environment where agents operate real Android applications through system-level control interfaces. This closed-loop setting captures dynamic execution conditions such as runtime delays, UI transitions, and system responses, better reflecting practical mobile assistant scenarios where interaction efficiency and system responsiveness are critical. 
Table~\ref{tab:androidworld} summarizes the key characteristics of the AndroidWorld benchmark. 
In addition to the benchmark evaluation, we also conduct an end-to-end real-world case study on a physical smartphone to validate the practicality of \name in real deployment scenarios.

\noindent\textbf{Devices.}
We evaluate \name on three representative on-device platforms (as shwon in Table~\ref{tab:devices}, including a Samsung Galaxy S24 smartphone, an NVIDIA Jetson AGX Orin, and a MacBook Air M4 laptop. These platforms represent typical deployment environments for mobile GUI agents with different compute capabilities.

\begin{table}[t]
    \centering
    \begin{tabular}{lccp{4.5cm}}
    \toprule
    \textbf{Difficulty} & \textbf{Count} & \textbf{Ratio} & \textbf{Example of Task} \\
    \midrule
    Easy   & 61 & 52.6\% & AudioRecorderRecordAudio \\
    Medium & 36 & 31.0\% & NotesTodoItemCount \\
    Hard   & 19 & 16.4\% & ExpenseAddMultipleFromMarkor \\
    \bottomrule
    \end{tabular}
    \caption{Task summary of AndroidWorld~\cite{rawles2024androidworld}.}
    \label{tab:androidworld}
\end{table}

\noindent\textbf{Implementation.}
\name is implemented as an end-to-end mobile GUI agent system. During evaluation, AndroidWorld tasks are executed in an Android emulator on a host machine, which runs the interaction environment and provides GUI states (e.g., screenshots and accessibility trees). The agent communicates with the model through HTTP requests. To emulate practical on-device deployment while ensuring stable latency measurement, we adopt a dual-device setup: the emulator executes UI interactions, while model inference runs on target hardware platforms (phone, NVIDIA Jetson, and laptop). The models are deployed using \texttt{llama.cpp} with Q8 quantization and served through the \texttt{vLLM} framework, providing a unified API interface for the AndroidWorld agent. On the smartphone, \texttt{llama.cpp} runs inside the Termux application to enable local model execution. This design allows us to simulate realistic on-device agent execution while measuring the true end-to-end latency of the perception–reasoning–action loop across heterogeneous devices. The evaluation are mainly based on a 4B VLM ~\cite{yan2025step}.

\begin{table}
    \centering
    \begin{tabular}{lcc}
    \toprule
    Device & Compute Units & Memory \\
    \midrule
    Samsung Galaxy S24 & CPU, GPU, NPU & 12\,GB \\
    Jetson AGX Orin & CPU, GPU & 64\,GB \\
    MacBook Air M4 & CPU, GPU & 24\,GB \\
    \bottomrule
    \end{tabular}
    \caption{Mobile and edge devices used in evaluation.}
    \label{tab:devices}
\end{table}

\noindent\textbf{Baselines.} We compare \name with the following representative baselines based on 4B VLMs/LLMs.
\begin{itemize}[leftmargin=12pt]
    \item \textbf{M3A Agent~\cite{rawles2024androidworld}}: Base mobile GUI agent that uses a VLM to reason over the current screen in a sequential perception–reasoning–action loop.
    
    \item \textbf{T3A Agent~\cite{rawles2024androidworld}}: A mobile GUI agent that relies on accessibility trees and performs step-wise reasoning using an LLM.

    \item \textbf{Input-pruning VLM agent}: Agent that reduces visual token overhead via screenshot or token pruning~\cite{lin2025showui}.

    \item \textbf{Offline exploration agent}: Agent that construct knowledge through offline exploration~\cite{zhao2025llm, xie2025gui, zhang2025webpilot}.


\end{itemize}

\noindent Beyond the above baselines, we further compare our method with existing methods reported on the AndroidWorld~\cite{rawles2307android} leaderboard.

\noindent \textbf{Evaluation metrics}. 
We evaluate \name in terms of both effectiveness and system efficiency on real on-device execution. \textbf{Success rate} measures the percentage of tasks successfully completed within a predefined step budget, indicating the overall task-solving capability of the agent. \textbf{Total steps} denotes the number of interaction steps required to finish a task, reflecting decision efficiency and the amount of trial-and-error during long-horizon execution. \textbf{Latency} is evaluated at two granularities. Step latency is defined as the time of a single interaction cycle, measured from capturing the current screenshot $s_i$ to completing the execution of the selected action $a_i$. This includes perception, model reasoning, online exploration, and action execution. 
End-to-end latency is defined as the total task completion time from task initialization to the final successful state, i.e., the sum of all step latencies across the task.
We also measure overhead like CPU usage, memory and power consumption in Section~\ref{subsec:overhead}.

\subsection{Overall Performance}

\begin{figure}
    \centering
    \setlength{\abovecaptionskip}{0.cm}
    \setlength{\belowcaptionskip}{-0.cm}
    \begin{subfigure}[t]{\linewidth}
        \centering
    \setlength{\abovecaptionskip}{0.cm}
    \setlength{\belowcaptionskip}{-0.cm}
        \includegraphics[width=\linewidth]{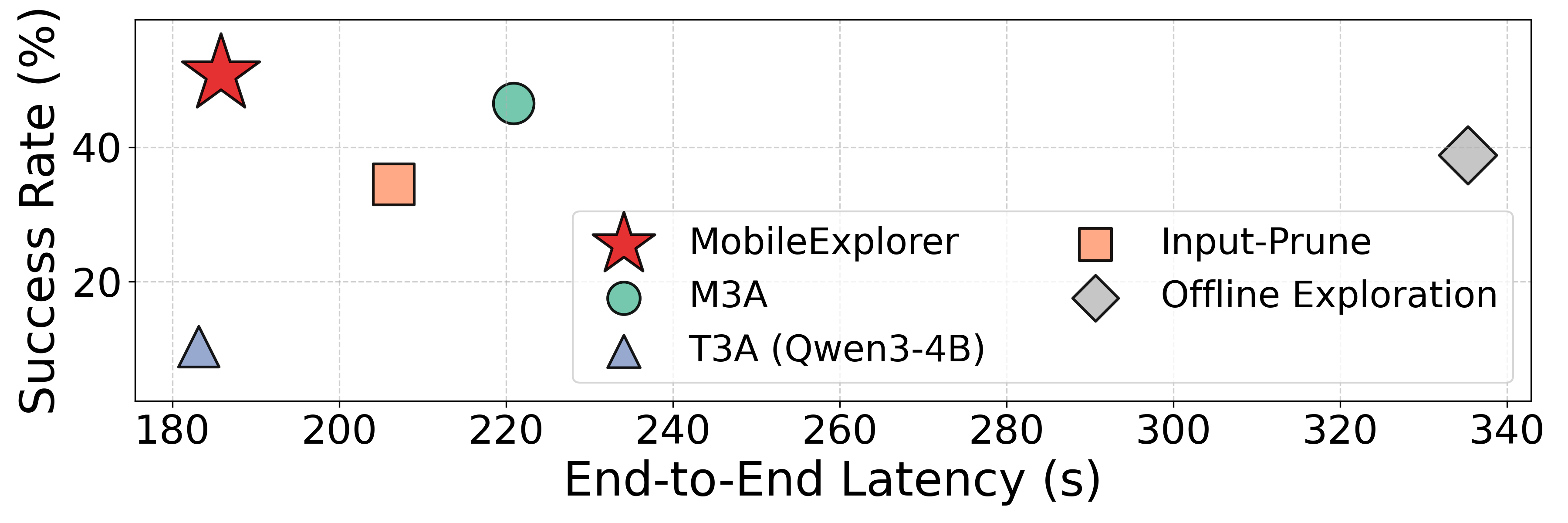}
        \caption{Success rate and end-to-end latency.}
        \label{fig:overall performance}    
    \end{subfigure}\hfill
    \begin{subfigure}[t]{\linewidth}
        \centering
    \setlength{\abovecaptionskip}{0.cm}
    \setlength{\belowcaptionskip}{-0.cm}
        \includegraphics[width=\linewidth]{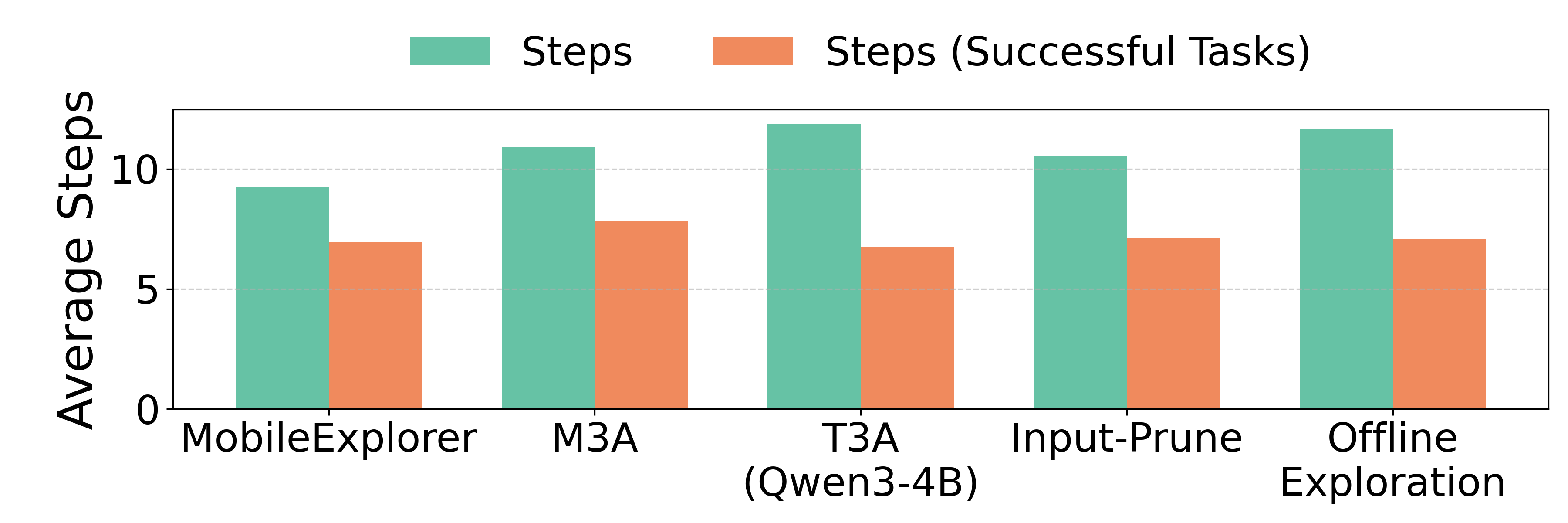}
        \caption{Average reasoning steps.}
        \label{fig:overall steps}
    \end{subfigure}
    \caption{Overall performance on AndroidWorld.}
    \label{fig:overall}
\end{figure}

\begin{table}
    \centering
    \setlength{\abovecaptionskip}{0.cm}
    \setlength{\belowcaptionskip}{-0.cm}
    \begin{tabular}{lc}
    \toprule
    \textbf{Method} & \textbf{AndroidWorld (\%)} \\
    \midrule
    MobileGPT~\cite{lee2024mobilegpt} & 23.0 \\
    AutoDroid-V2~\cite{wen2025autodroid} & 26.0 \\
    M3A (a11y, GPT-4-Turbo)~\cite{rawles2024androidworld} & 30.6 \\
    M3A (a11y, Gemini-2.5-Pro)~\cite{rawles2024androidworld} & 31.0 \\
    M3A (SoM, GPT-4-Turbo)~\cite{rawles2024androidworld} & 25.4 \\
    M3A (SoM, Gemini-2.5-Pro)~\cite{rawles2024androidworld} & 39.7 \\
    GLM-4.1V-9B-Thinking~\cite{hong2025glm} & 41.7 \\
    UI-TARS (UI-TARS-7B)~\cite{wang2025ui} & 33.0 \\
    \midrule
    \textbf{MobileExplorer} & \textbf{50.9} \\
    \bottomrule
    \end{tabular}
        \caption{Success Rate (\%) on AndroidWorld.}
    \label{tab:success rate}
\end{table}

We evaluate \name on the AndroidWorld benchmark under a fully on-device setting. 
As shown in Table~\ref{tab:success rate} and Fig.~\ref{fig:overall performance}, \name achieves a success rate of 50.86\% (59/116 tasks), which is the best among all compared methods. 
Compared with the VLM-based baseline M3A, which achieves 46.55\%, this corresponds to a 9.3\% relative improvement. 
This improvement indicates that integrating latency-bounded online exploration into the reasoning process allows the agent to obtain more useful UI context and make better interaction decisions. 
As a result, \name improves the ability of mobile GUI agents to complete long-horizon tasks under the on-device setting.

Meanwhile, \name also improves the efficiency of task execution. 
As illustrated in Fig.~\ref{fig:overall steps}, our method requires only 9.24 interaction steps on average, compared with 10.93 steps for M3A, corresponding to a 15.5\% reduction in interaction cost. 
This reduction in trial-and-error operations further translates into lower end-to-end latency. 
In particular, \name completes tasks in 185.82\,s on average, reducing the overall latency by 15.9\% compared with M3A. 
Importantly, this improvement is achieved without increasing per-step reasoning overhead, since exploration is performed in parallel with model reasoning and utilizes otherwise idle inference time.

\subsection{Real-world Case Study}

\begin{figure}
    \centering
    \setlength{\abovecaptionskip}{0.cm}
    \setlength{\belowcaptionskip}{0.cm}
    \includegraphics[width=1.0\linewidth]{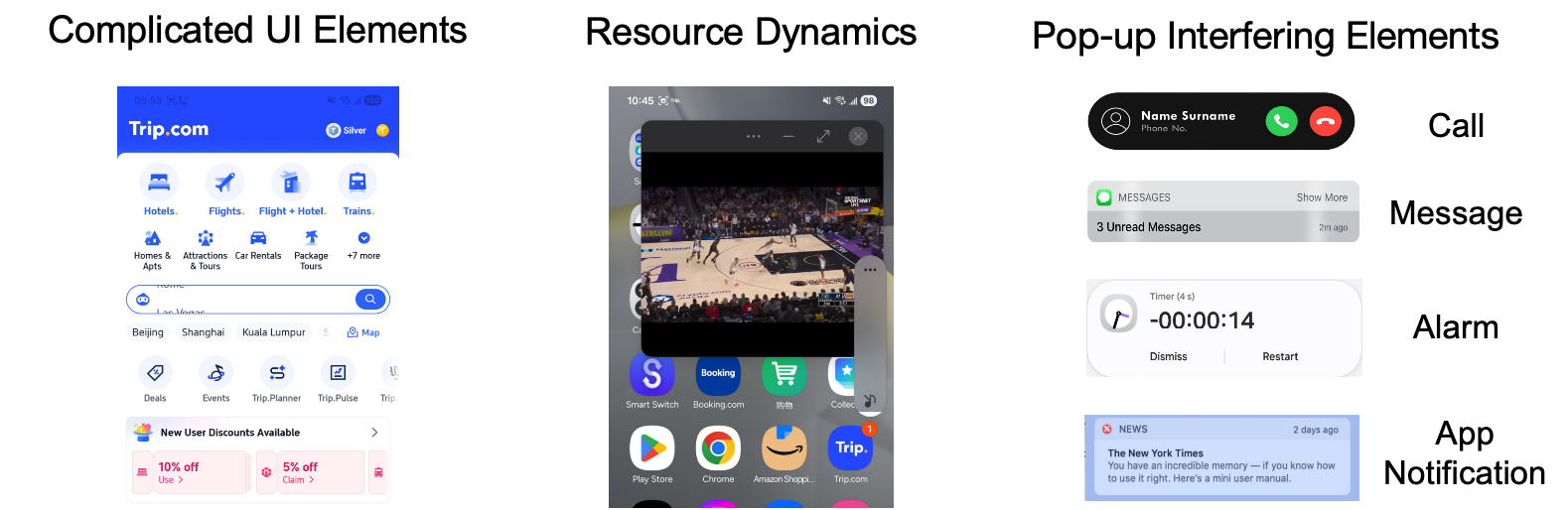}
    \caption{Examples of settings in the case study.}
    \label{fig:case study setting}
\end{figure}

While AndroidWorld provides a controlled benchmark for evaluating mobile GUI agents, its tasks are relatively simplified and do not capture several important characteristics of real mobile environments. In practice, mobile applications contain significantly more complex UI structures, experience dynamic resource conditions, and frequently introduce interrupt-driven UI changes such as system dialogs or notifications. To evaluate how these factors affect agent behavior, we construct a set of real-world smartphone tasks using popular mobile applications and analyze them along three practical dimensions, as shwon in Fig.~\ref{fig:case study setting}.

\begin{table}
    \centering
    \begin{tabular}{p{2cm}p{3cm}p{2.5cm}}
    \toprule
    \textbf{Setting} & \textbf{Feature} & \textbf{Task} \\
    \midrule
    Complicated UI Elements & \textbf{48 elements/page} in average (while <20 in AndroidWorld) & Trip planning for a city in Trip APP. \\
    \hline
    Pop-up Interfering Elements & Type of Interfering: \textbf{alarm}, \textbf{message}, \textbf{call}, \textbf{app notifications}  & Input text in Notes APP. \\
    \hline
    Resource Dynamics & Background tasks: \textbf{video and music playing} & Modify system settings (Bluetooth, WiFi). \\
    \bottomrule
    \end{tabular}
    \caption{Task categories and features in the case study.}
    \label{tab:devices}
\end{table}

\noindent\textbf{Complicated UI Elements.}
Many AndroidWorld tasks are executed in relatively simple applications with limited UI elements and shallow interface hierarchies. On average, AndroidWorld pages contain about 19.75 interactive elements, whereas real-world mobile applications often present much denser interfaces. For instance, typical screens in the Trip application contain around 48 interactive elements due to long result lists and recommendation panels. To evaluate the agent under such conditions, we design a trip planning task where the agent searches for attractions in the Trip application. We randomly generate 10 different city queries to create diverse UI states, requiring the agent to navigate through dense search results to locate the target item.

\noindent\textbf{Interrupt On-pop Elements.}
Real mobile environments often introduce unexpected interruptions that temporarily modify the interface structure or occlude interactive elements, such as alarms, message notifications, or incoming calls. These interruptions are largely absent in AndroidWorld, where tasks run in stable environments. To emulate such scenarios, we design a stopwatch task in which different types of interruptions (alarm, message, call, and application notifications) are injected during execution, forcing the agent to correctly recover the UI state before continuing.

\begin{figure}
    \centering
    \setlength{\abovecaptionskip}{0.cm}
    \setlength{\belowcaptionskip}{0.cm}
    \includegraphics[width=1.0\linewidth]{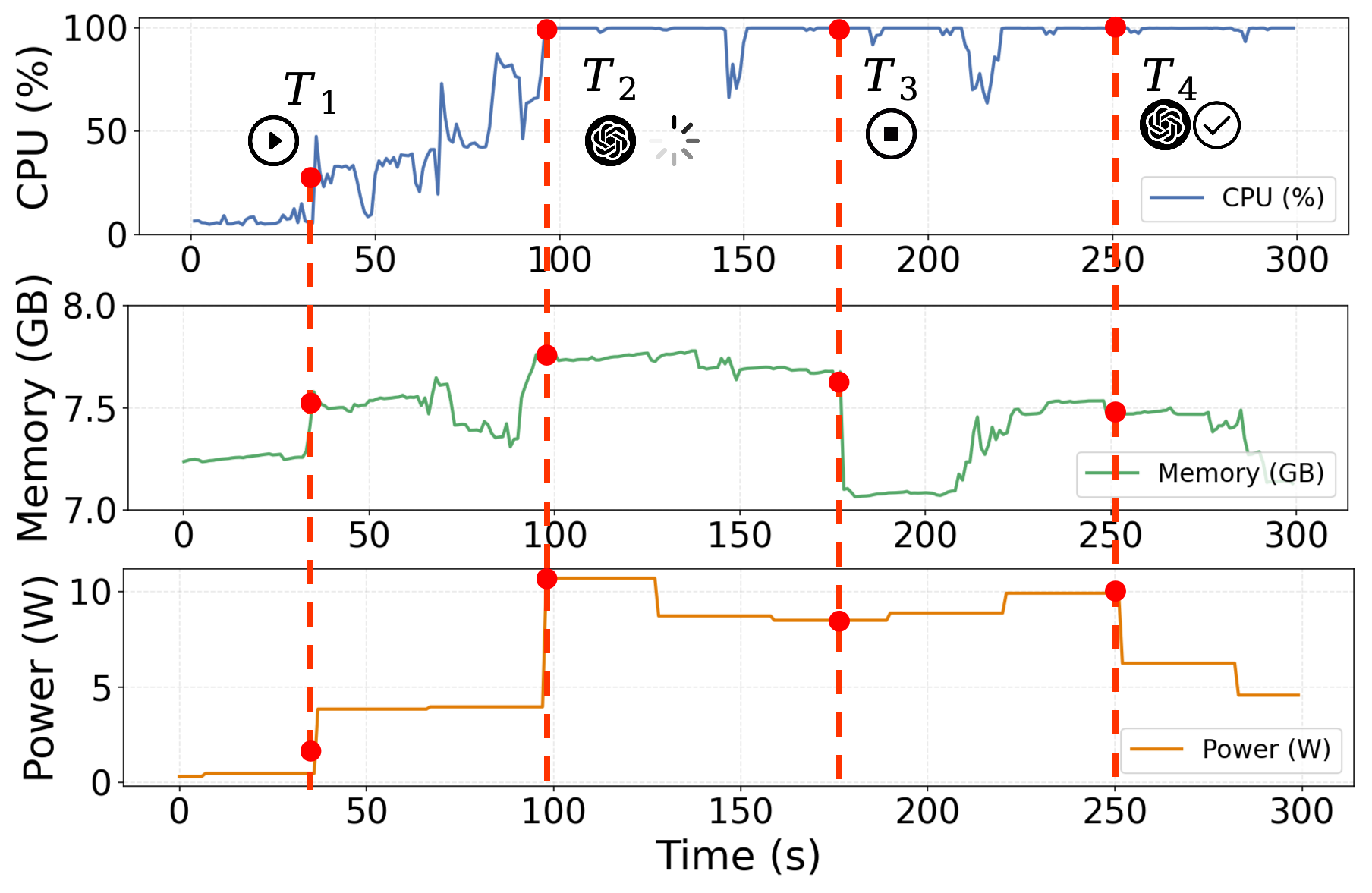}
    \caption{Resource dynamics on mobile phone. $T_1$ to $T_3$: Video playing, $T_2$ to $T_4$: VLM reasoning.}
    \label{fig:resource dynamic}
\end{figure}

\noindent\textbf{Resource Dynamics.}
Unlike controlled benchmark environments, real smartphones operate under dynamic system conditions where available resources fluctuate due to concurrent background activities, as shown in Fig.~\ref{fig:resource dynamic}. Background workloads such as video playback or music streaming may consume CPU and memory resources, affecting the latency of on-device perception and reasoning modules. Since AndroidWorld does not model such resource contention, we design a system settings task where the agent toggles Bluetooth or WiFi while background applications run concurrently, creating realistic resource dynamics during execution.

\begin{figure}
    \raggedright   
    \setlength{\abovecaptionskip}{0.cm}
    \setlength{\belowcaptionskip}{-0.cm}
    \begin{subfigure}{0.49\linewidth}
            \setlength{\abovecaptionskip}{0.cm}
        \setlength{\belowcaptionskip}{0.cm}
        \raggedright   
        \includegraphics[width=1\linewidth]{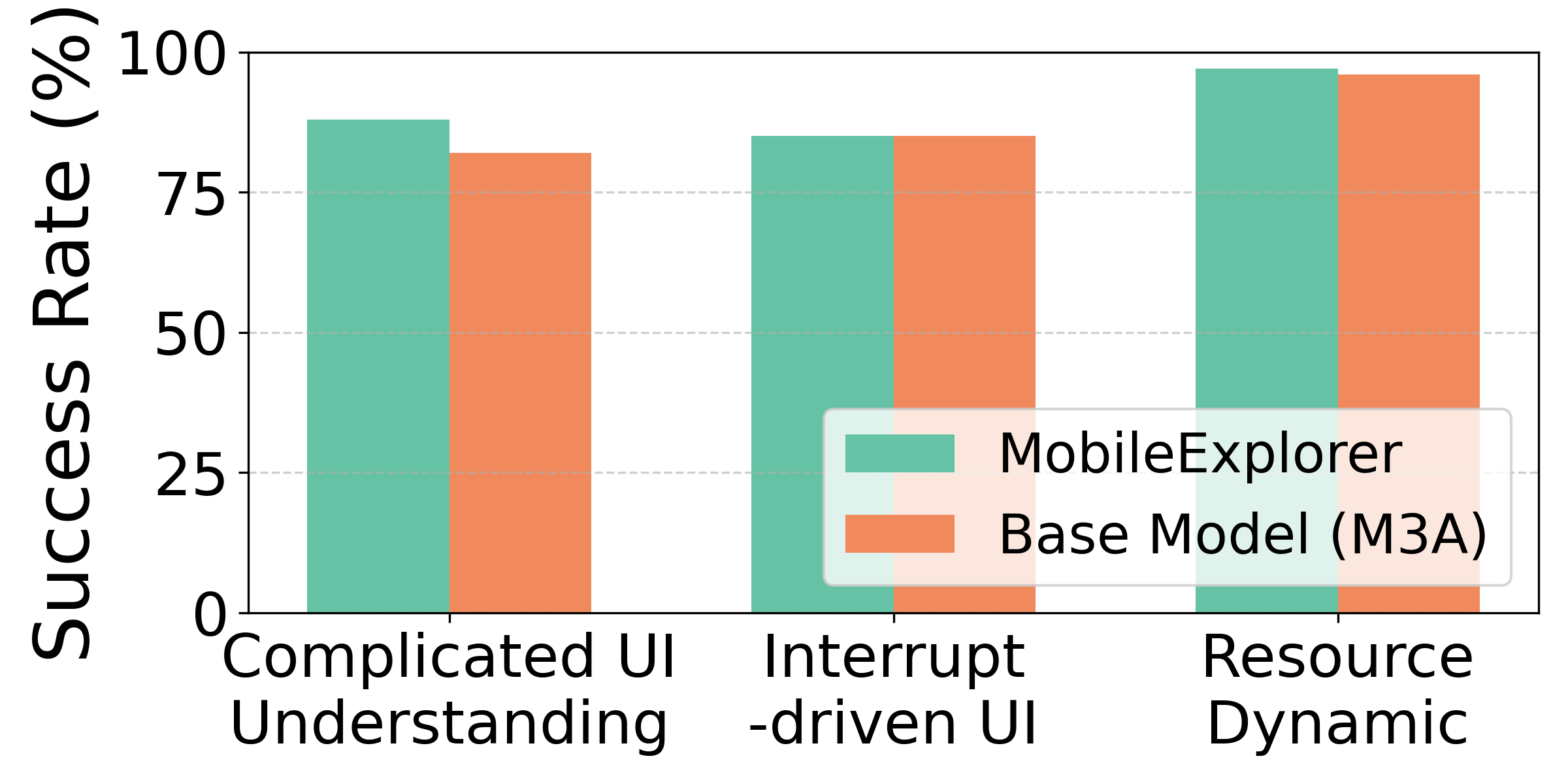}
        \caption{Success rate.}
        \label{fig:case study accuracy}
    \end{subfigure}%
    \hspace{0.5mm}
    \begin{subfigure}{0.49\linewidth}
            \setlength{\abovecaptionskip}{0.cm}
        \setlength{\belowcaptionskip}{0.cm}
        \raggedright   
        \includegraphics[width=1\linewidth]{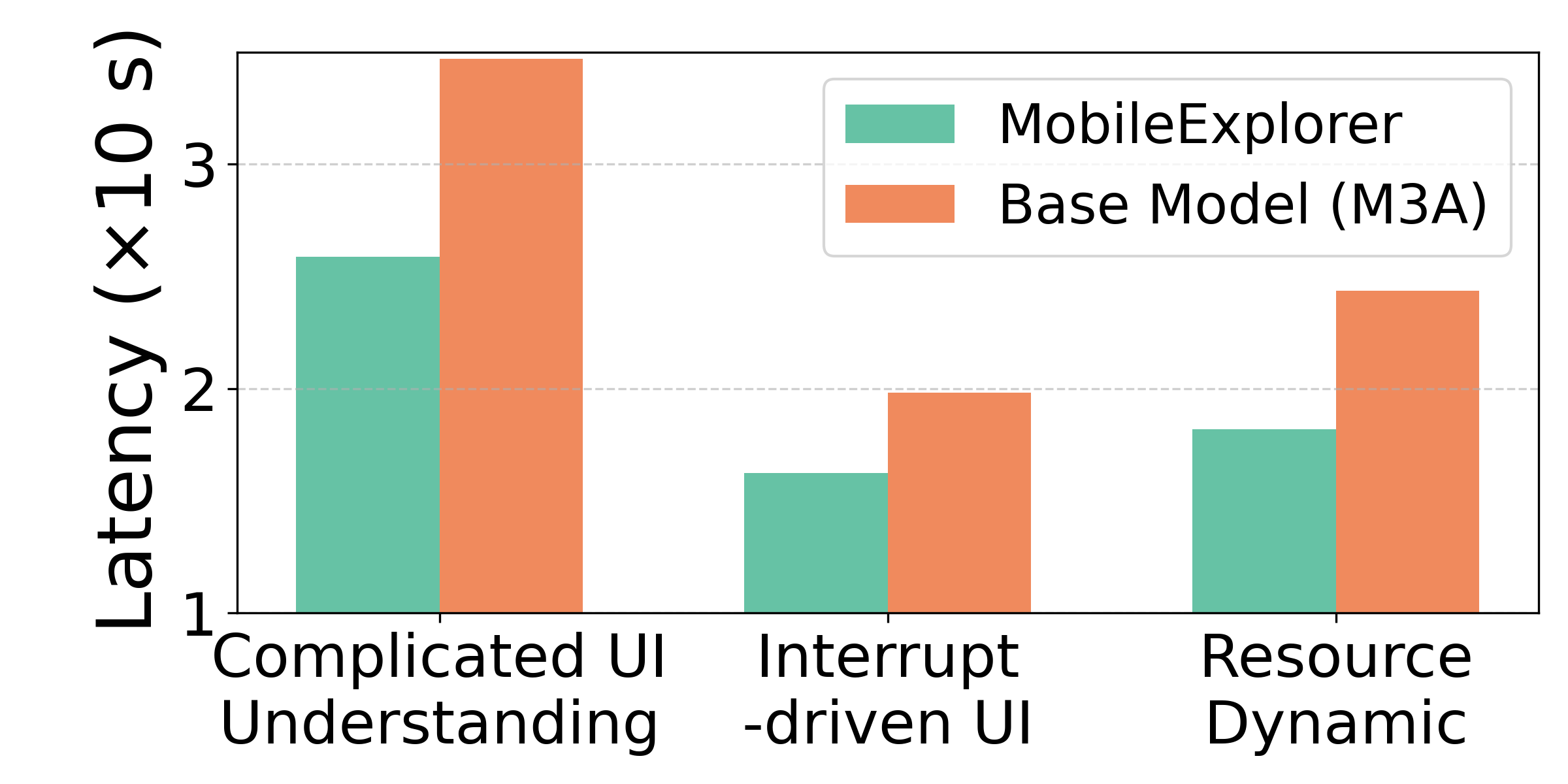}
        \caption{End-to-End latency.}
        \label{fig:case study latency}
    \end{subfigure}%
    \caption{Performance in different complicated settings of the case study.}
    \label{fig:case study performance}
\end{figure}

\noindent\textbf{Results.}
Together, these tasks evaluate \name in realistic mobile scenarios beyond the AndroidWorld benchmark. Each task is repeated three times and the averaged results are shown in Fig.~\ref{fig:case study performance}. \name consistently reduces end-to-end latency while achieving comparable or higher success rates across all settings. The improvement is most significant for complicated UI environments, and the results also demonstrate robustness under interrupt-driven UI changes and dynamic resource conditions.

\subsection{Understanding \name's Performance}

\subsubsection{Ablation Study}
Fig.~\ref{fig:ablation study} evaluates the contribution of key components in \name. 
Replacing the task-relevance-driven element selection with random exploration reduces the success rate from 50.9\% to 42.2\%, as random exploration often probes task-irrelevant UI elements within the limited reasoning window. 
Removing the two-level rollback mechanism causes the largest degradation, reducing the success rate to 39.7\% and increasing end-to-end latency because exploration may fail to return to the original UI state. 
Disabling exploration alignment further decreases the success rate to 47.4\%, since unfiltered exploration observations may introduce misleading information into the prompt. 
These results highlight the importance of task-aware exploration, reliable state recovery, and exploration–reasoning alignment.

\subsubsection{Performance on different task categories}
\begin{figure}
    \raggedright   
    \setlength{\abovecaptionskip}{0.cm}
    \setlength{\belowcaptionskip}{-0.cm}
    \begin{subfigure}{0.49\linewidth}
            \setlength{\abovecaptionskip}{0.cm}
        \setlength{\belowcaptionskip}{0.cm}
        \raggedright   
        \includegraphics[width=1\linewidth]{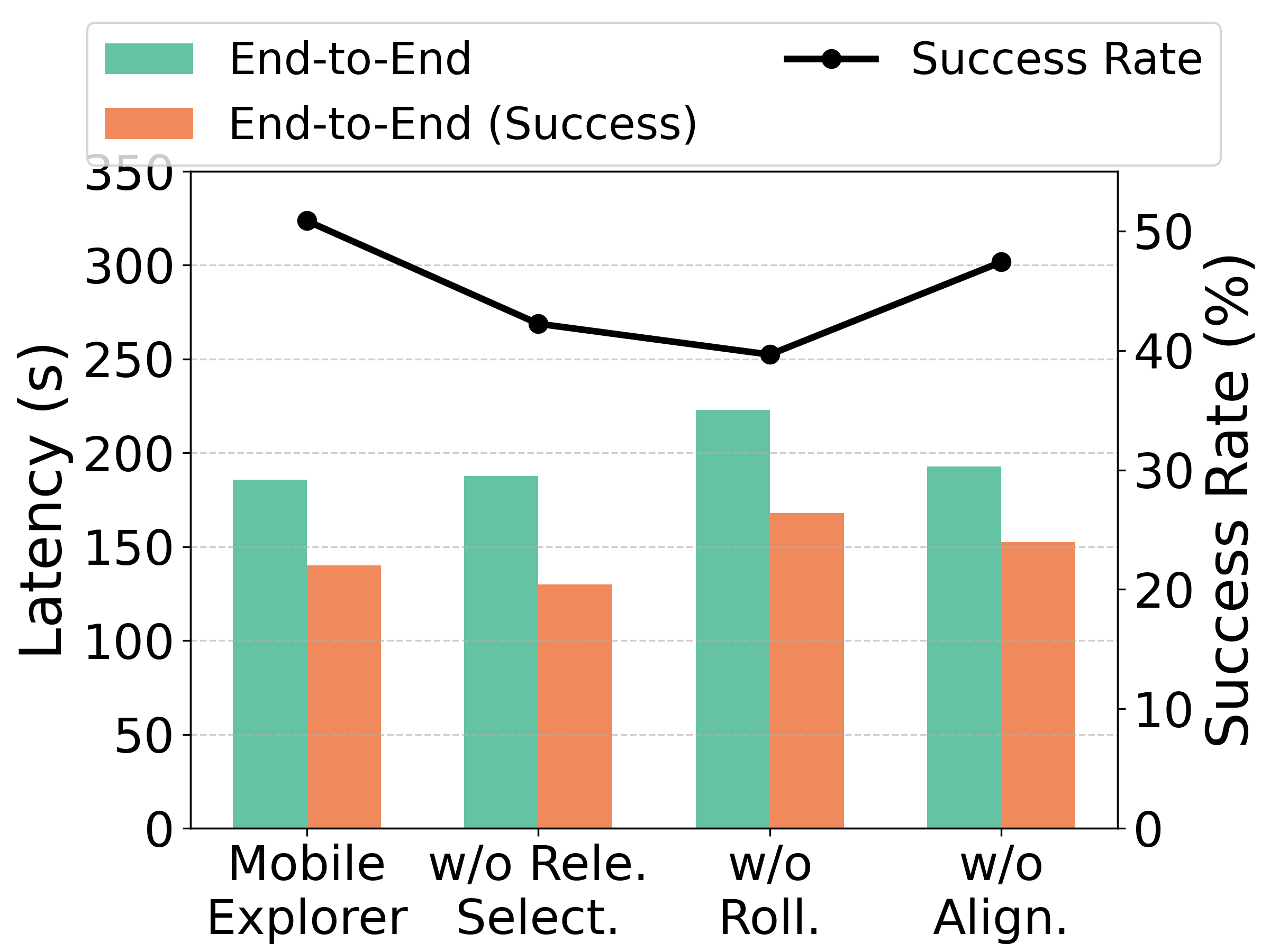}
        \caption{Ablation study.}
        \label{fig:ablation study}
    \end{subfigure}%
    \hspace{0.5mm}
    \begin{subfigure}{0.49\linewidth}
            \setlength{\abovecaptionskip}{0.cm}
        \setlength{\belowcaptionskip}{0.cm}
        \raggedright   
        \includegraphics[width=1\linewidth]{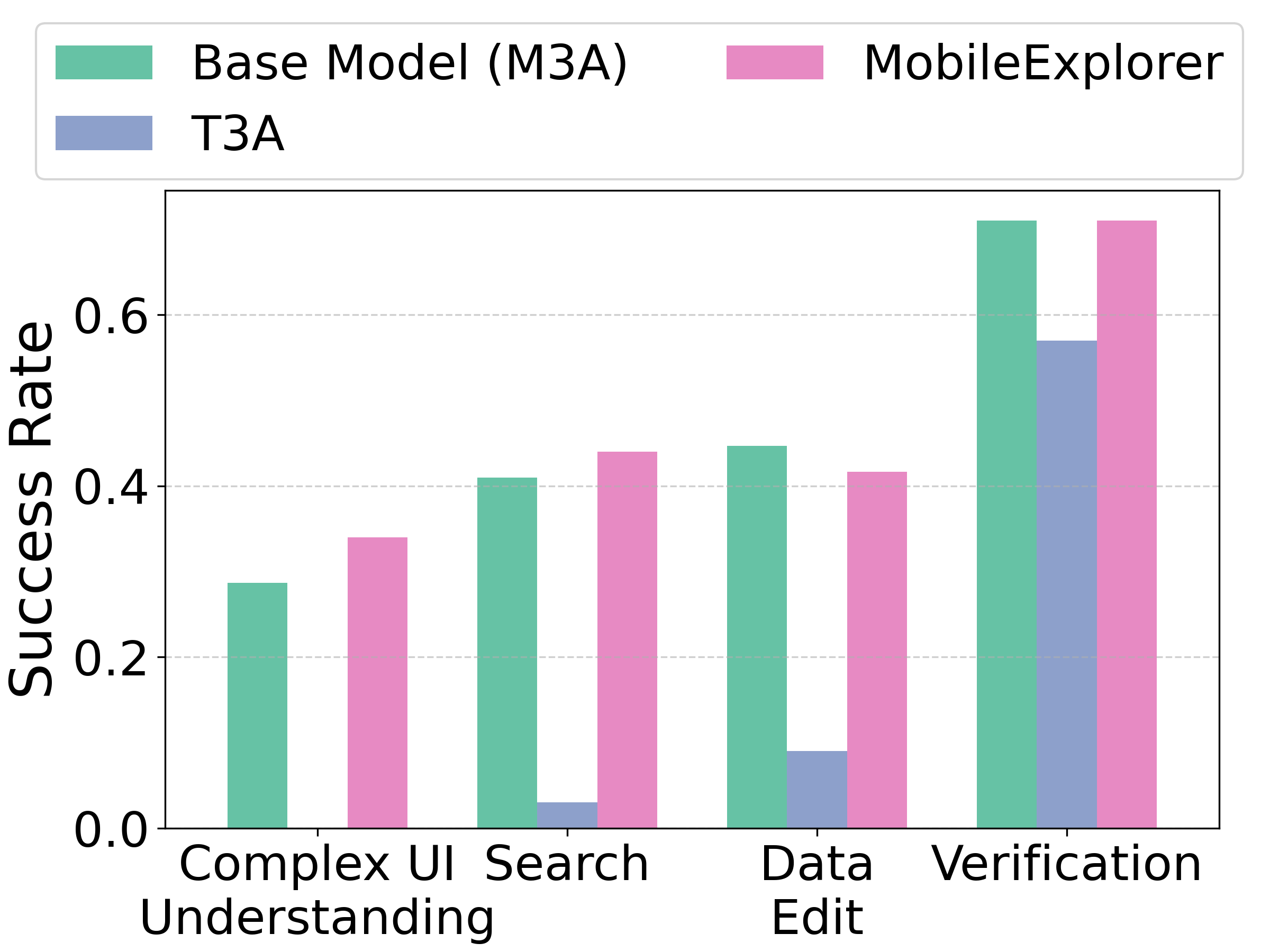}
        \caption{Task categories.}
        \label{fig:task types}
    \end{subfigure}%
    \hspace{0.5mm}
    \begin{subfigure}{0.49\linewidth}
            \setlength{\abovecaptionskip}{0.cm}
        \setlength{\belowcaptionskip}{0.cm}
        \raggedright   
        \includegraphics[width=1\linewidth]{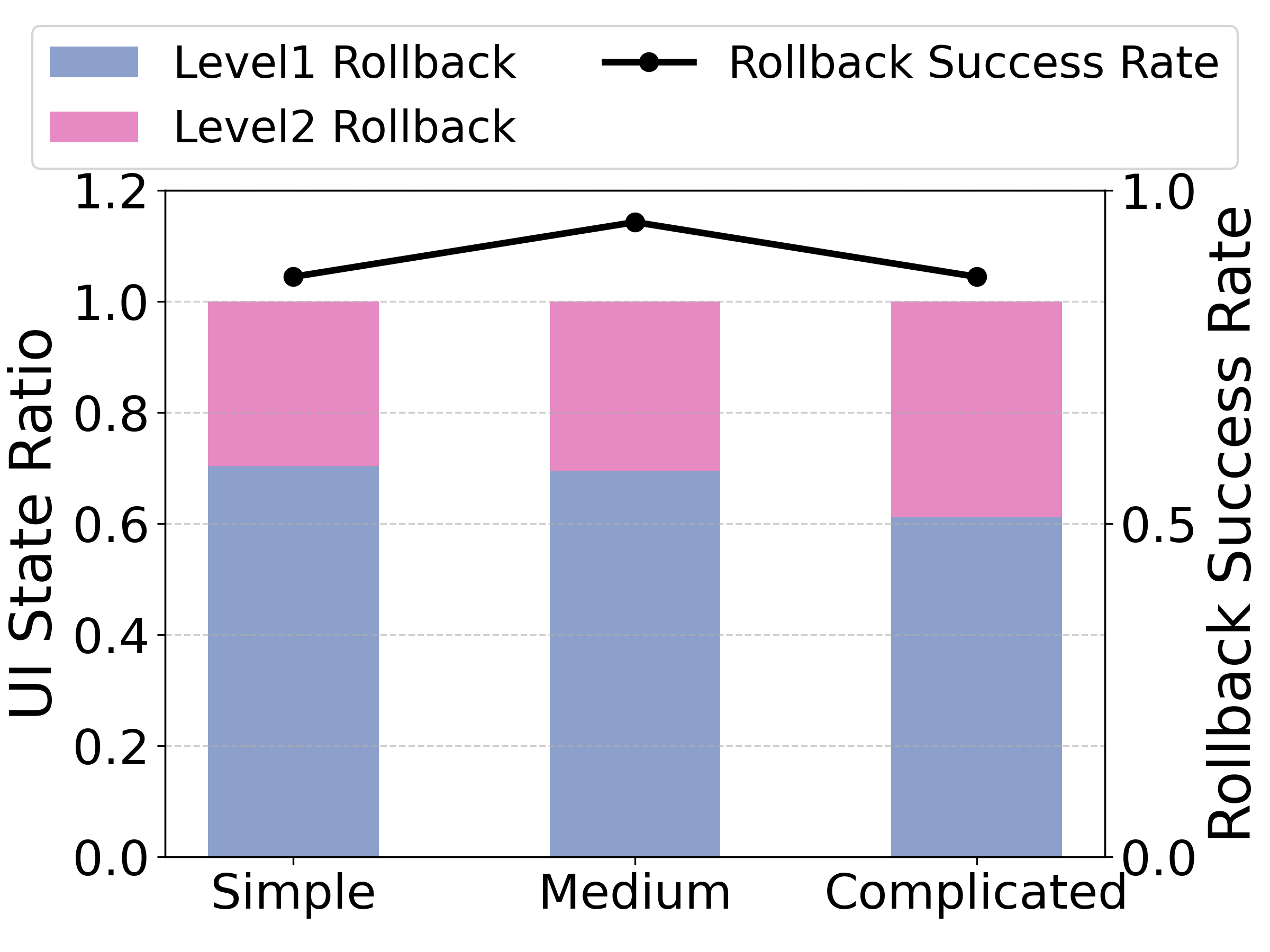}
        \caption{Effectiveness of two-level rollback strategy.}
        \label{fig:rollback ratio}
    \end{subfigure}%
    \hspace{0.5mm}
    \begin{subfigure}{0.49\linewidth}
            \setlength{\abovecaptionskip}{0.cm}
        \setlength{\belowcaptionskip}{0.cm}
        \raggedright   
        \includegraphics[width=1\linewidth]{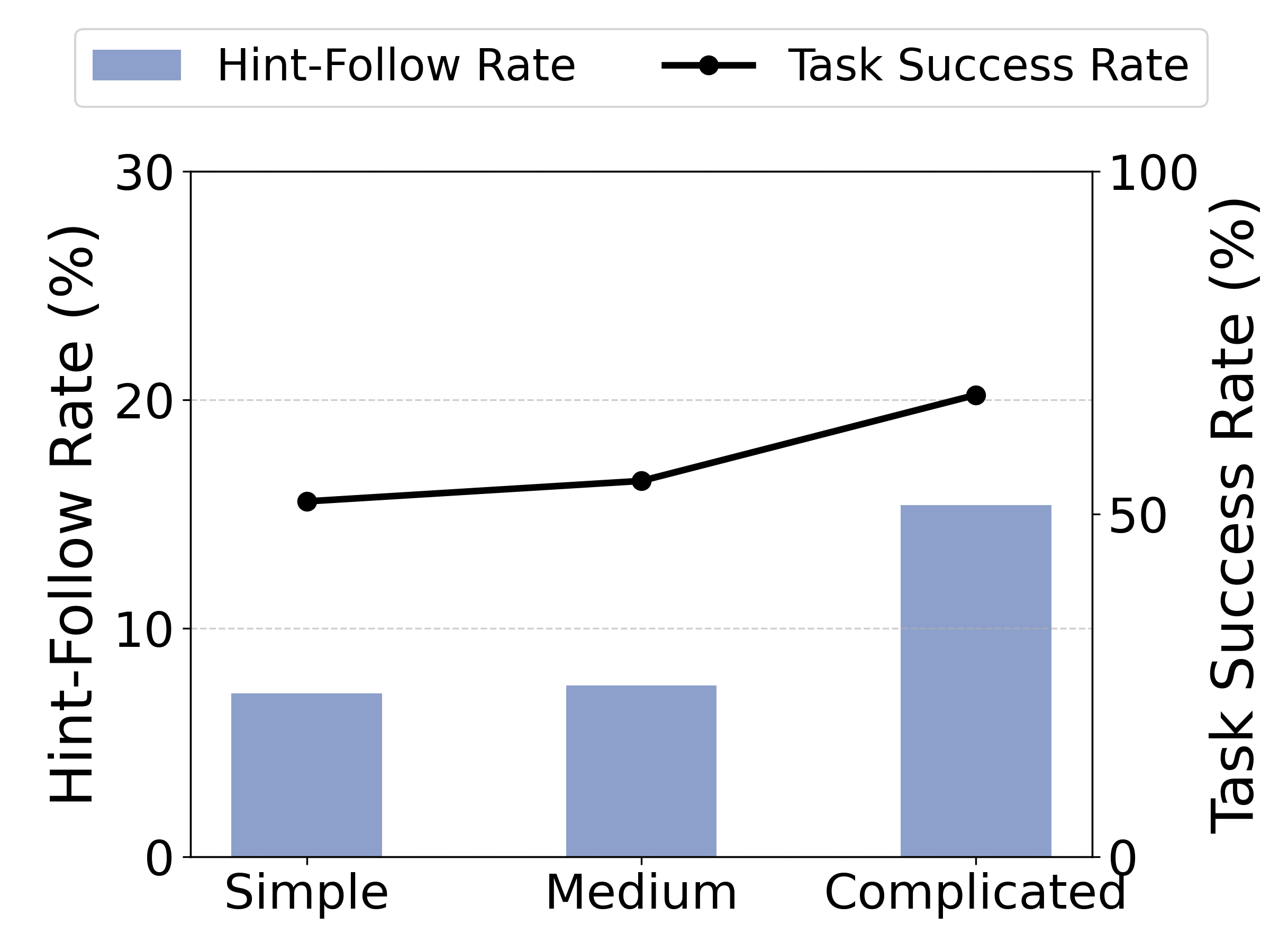}
        \caption{Effectiveness of exploration-augmented reasoning.}
        \label{fig:hint hit}
    \end{subfigure}%
    \caption{Understanding \name's performance. }
    \label{fig:understanding}
\end{figure}

AndroidWorld organizes tasks into several categories including \textit{complex UI understanding}, \textit{search}, \textit{information retrieval}, \textit{data entry}, \textit{data edit}, and \textit{verification}~\cite{rawles2024androidworld}. 
Fig.~\ref{fig:task types} shows that \name consistently improves performance on visually intensive tasks such as \textit{complex UI understanding}, \textit{search}, and \textit{information retrieval}, where identifying relevant UI elements among many candidates is critical. 
In contrast, for structured interaction tasks such as \textit{data entry} and \textit{data edit}, \name performs slightly below M3A because these tasks rely more on deterministic action sequences than UI exploration. 
Overall, \name is particularly effective for exploration-intensive tasks involving complex UI structures.

\subsubsection{Effectiveness of two-level rollback strategy.}
We analyze rollback events on UI pages categorized as \textit{Simple}, \textit{Medium}, and \textit{Complicated} based on the number of clickable elements. 
As shown in Fig.~\ref{fig:rollback ratio}, Level-1 rollback handles most cases, while complicated pages show a higher proportion of Level-2 rollbacks, indicating that simple backtracking is more likely to fail in deeper or less reversible navigation paths. 
Despite this, the overall rollback success rate remains consistently high across all categories, demonstrating that Level-2 recovery effectively complements Level-1 rollback.

\subsubsection{Effectiveness of exploration-augmented reasoning.}
To evaluate whether exploration-derived hints contribute to task completion, we measure the hint-follow rate and relate it to task success. 
We aggregate step-level hint hits into a task-level metric and group tasks into three complexity levels (\textit{Simple}, \textit{Medium}, and \textit{Complicated}). 
Fig.~\ref{fig:hint hit} shows that the hint-follow rate increases as UI complexity grows. 
Meanwhile, tasks with higher hint-follow rates consistently exhibit higher success rates across all complexity levels. 
This observation suggests that exploration-derived hints provide useful guidance, especially when the interface contains many candidate elements.

\subsection{Micro Benchmark Performance}

\begin{figure}
    \raggedright   
    \setlength{\abovecaptionskip}{0.cm}
    \setlength{\belowcaptionskip}{-0.cm}
    \begin{subfigure}{0.49\linewidth}
            \setlength{\abovecaptionskip}{0.cm}
        \setlength{\belowcaptionskip}{0.cm}
        \raggedright   
        \includegraphics[width=1\linewidth]{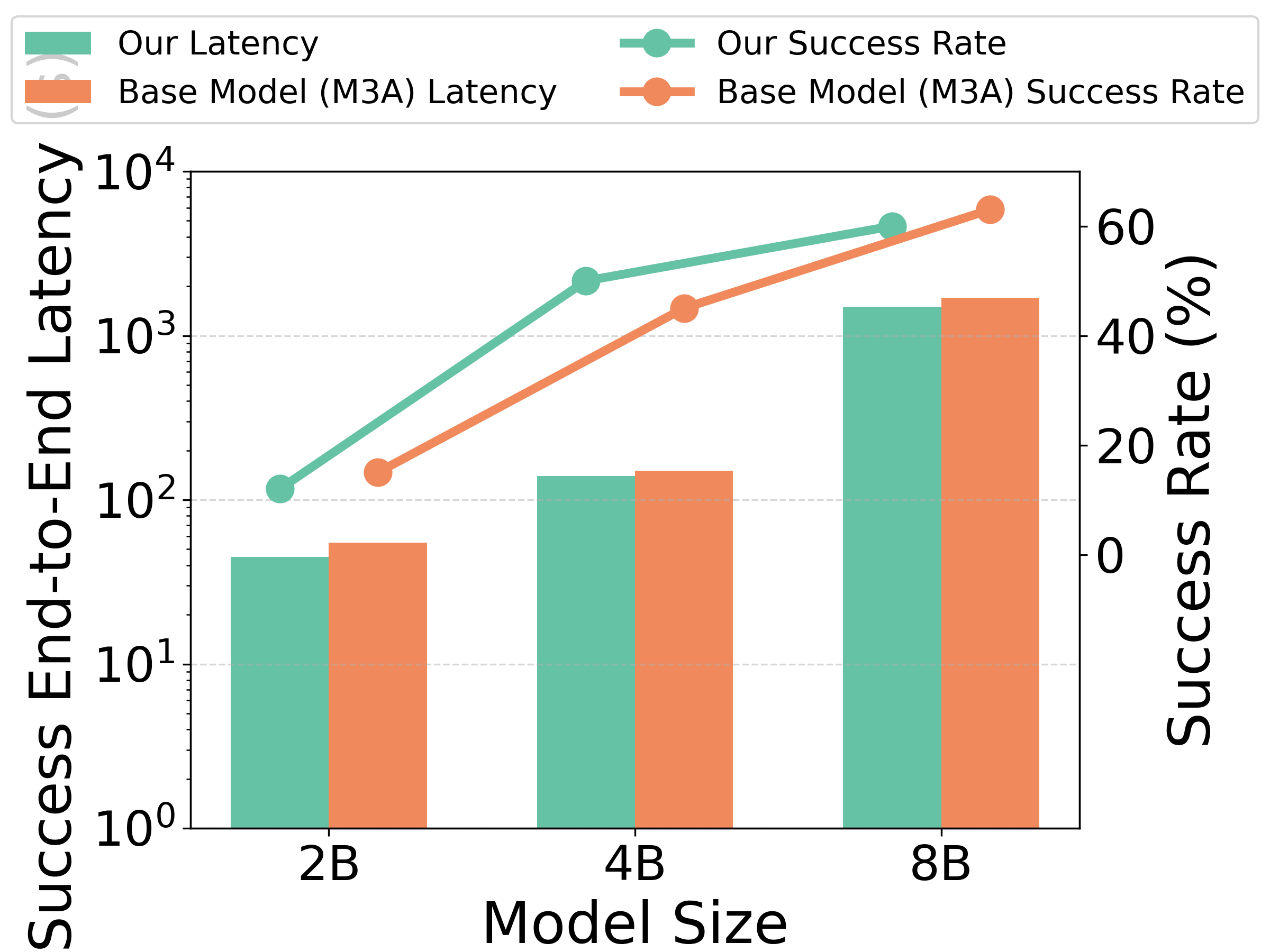}
        \caption{Different sizes of models.}
        \label{fig:model size}
    \end{subfigure}%
    \hspace{0.5mm}
    \begin{subfigure}{0.49\linewidth}
            \setlength{\abovecaptionskip}{0.cm}
        \setlength{\belowcaptionskip}{0.cm}
        \raggedright   
        \includegraphics[width=1\linewidth]{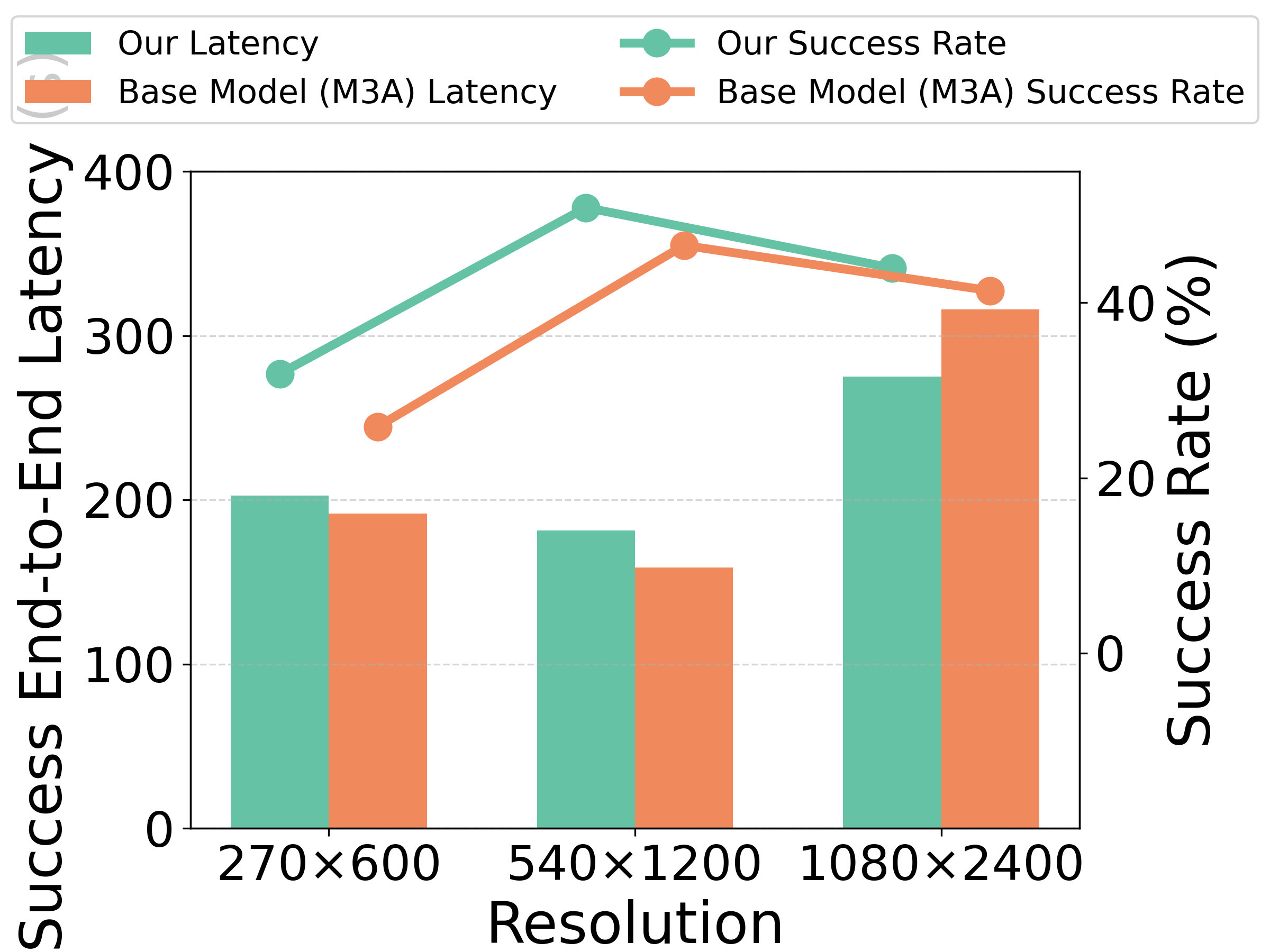}
        \caption{Different resolutions.}
        \label{fig:resolution}
    \end{subfigure}%
    \caption{Micro benchmark performance.}
    \label{fig:micro benchmark}
    \vspace{-1em}
\end{figure}

\subsubsection{Performance with different model sizes.}
To further examine the robustness of our design across different model scales, we evaluate the agents using MAI-UI models of different sizes, including 2B and 8B. Fig.~\ref{fig:model size} reports the end-to-end latency and success rate under these settings. Across different model sizes, \name\ consistently maintains comparable task success rates while reducing the average number of interaction steps, leading to lower end-to-end latency. In particular, our exploration-augmented design typically saves around one reasoning step per task, demonstrating that the benefit of online exploration generalizes across models with different capacities.

\subsubsection{Performance under different resolutions of screenshots.}
We evaluate three input resolutions to examine their impact on task performance. As shown in Fig.~\ref{fig:resolution}, higher resolutions generally improve success rates for both methods since more visual details of UI elements are preserved. MobileExplorer achieves its best performance at a moderate resolution, while M3A improves more gradually as resolution increases. Across all settings, MobileExplorer consistently requires fewer interaction steps than M3A, suggesting that exploration-based hints help guide the agent more efficiently. Overall, a moderate resolution provides a good balance between visual fidelity and interaction efficiency.

\subsection{System Overhead}
\label{subsec:overhead}

    
    
    
    
\begin{figure}
    \raggedright   
    \setlength{\abovecaptionskip}{0.cm}
    \setlength{\belowcaptionskip}{-0.cm}
    \begin{subfigure}{0.49\linewidth}
            \setlength{\abovecaptionskip}{0.cm}
        \setlength{\belowcaptionskip}{0.cm}
        \raggedright   
        \includegraphics[width=1\linewidth]{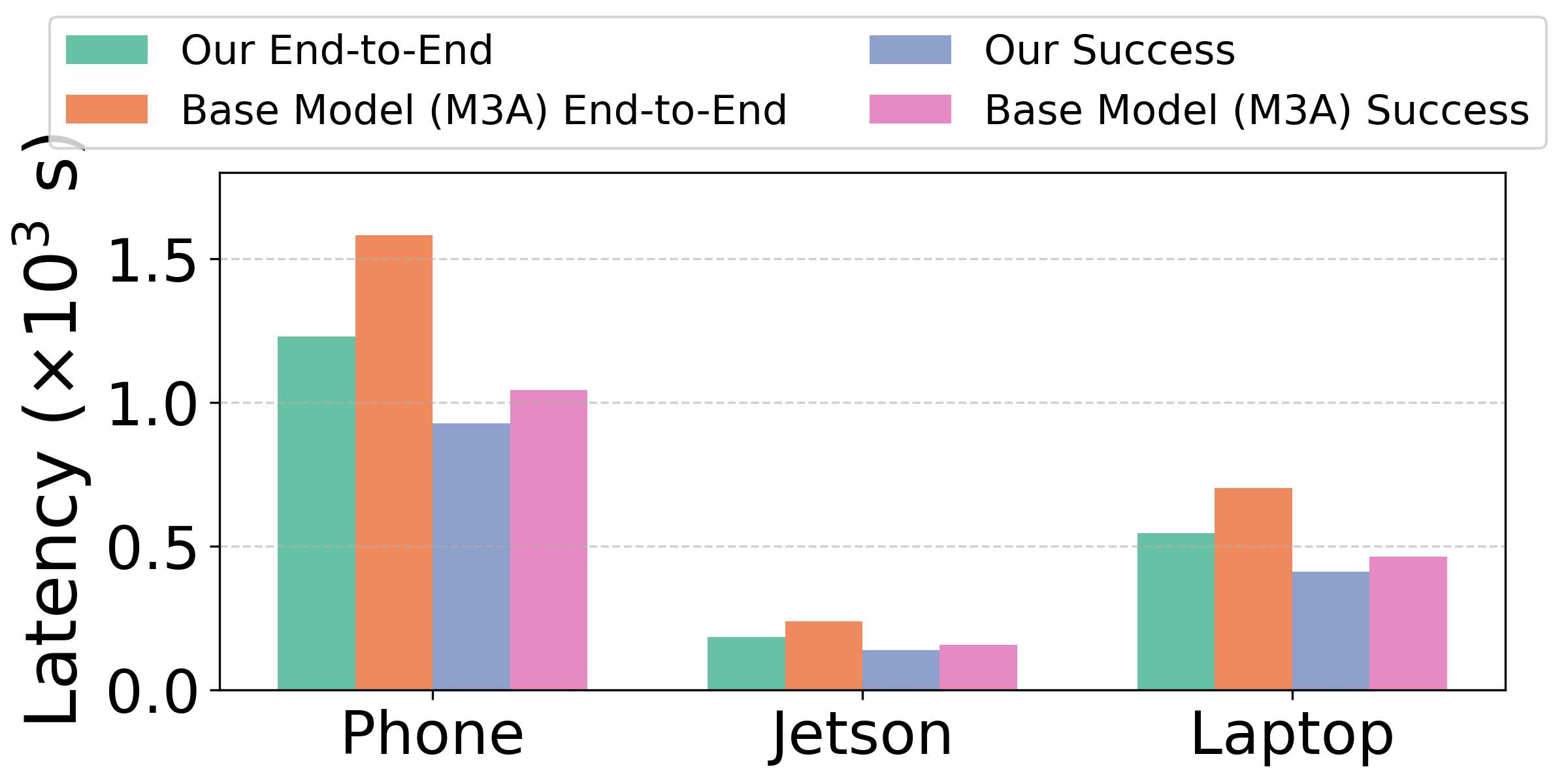}
        \caption{Latency.}
        \label{fig:platform}
    \end{subfigure}%
    \hspace{0.5mm}
    \begin{subfigure}{0.49\linewidth}
            \setlength{\abovecaptionskip}{0.cm}
        \setlength{\belowcaptionskip}{0.cm}
        \raggedright   
        \includegraphics[width=1\linewidth]{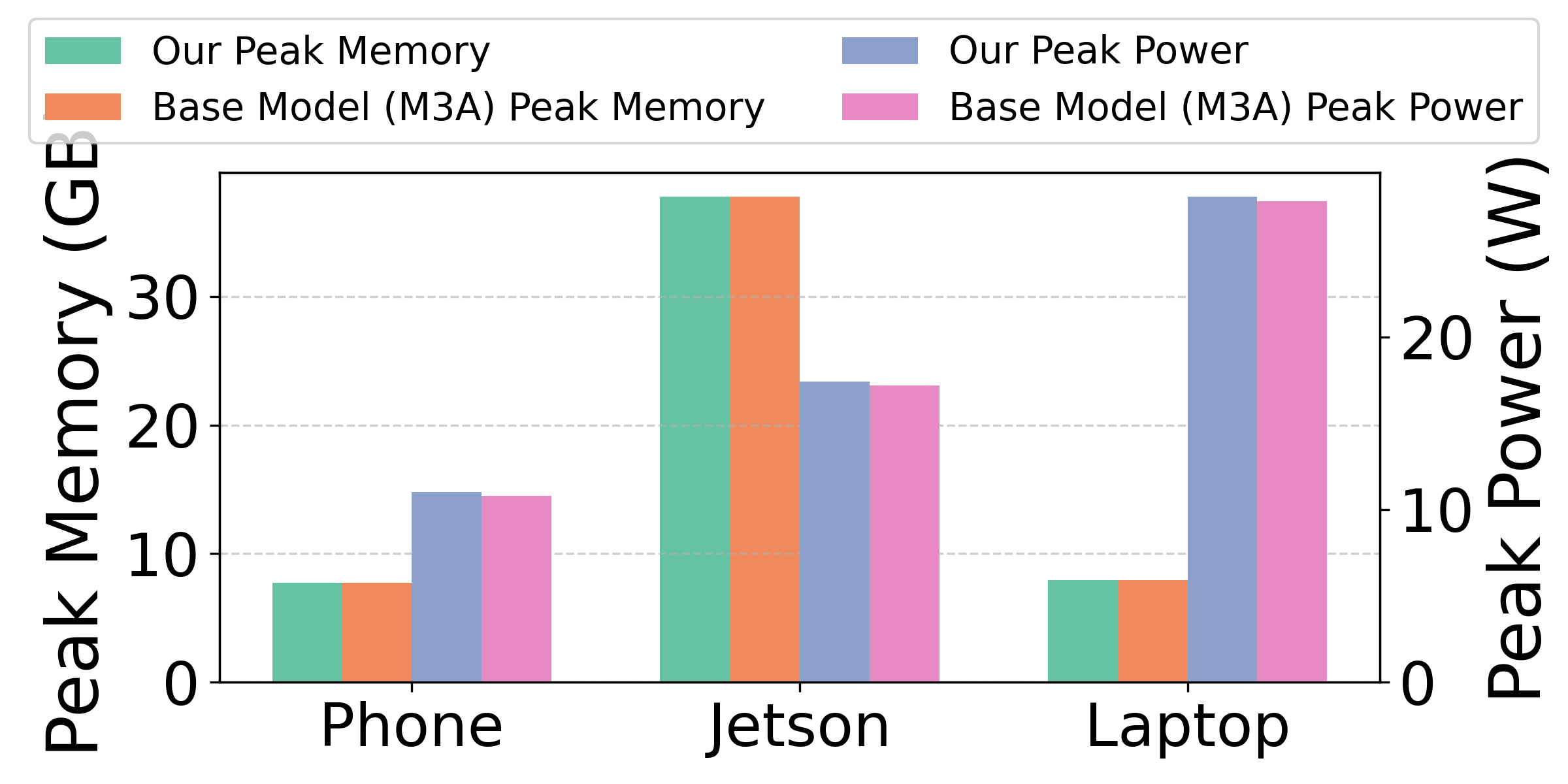}
        \caption{Memory and power.}
        \label{fig:overall overhead}
    \end{subfigure}%
    \caption{System overhead on different platforms.}
    \label{fig:overhead}
\end{figure}

We evaluate the system overhead of \name across different platforms, including a commercial smartphone, Jetson, and a laptop. Fig.~\ref{fig:platform} reports the end-to-end latency, peak memory usage, and peak power consumption. Across all devices, \name consistently reduces end-to-end latency compared with the baseline while maintaining nearly identical memory and power usage. This indicates that the exploration controller introduces only minimal additional system overhead.

\begin{figure}
    \raggedright   
    \setlength{\abovecaptionskip}{0.cm}
    \setlength{\belowcaptionskip}{-0.cm}
    \begin{subfigure}{0.33\linewidth}
            \setlength{\abovecaptionskip}{0.cm}
        \setlength{\belowcaptionskip}{0.cm}
        \raggedright   
        \includegraphics[width=1\linewidth]{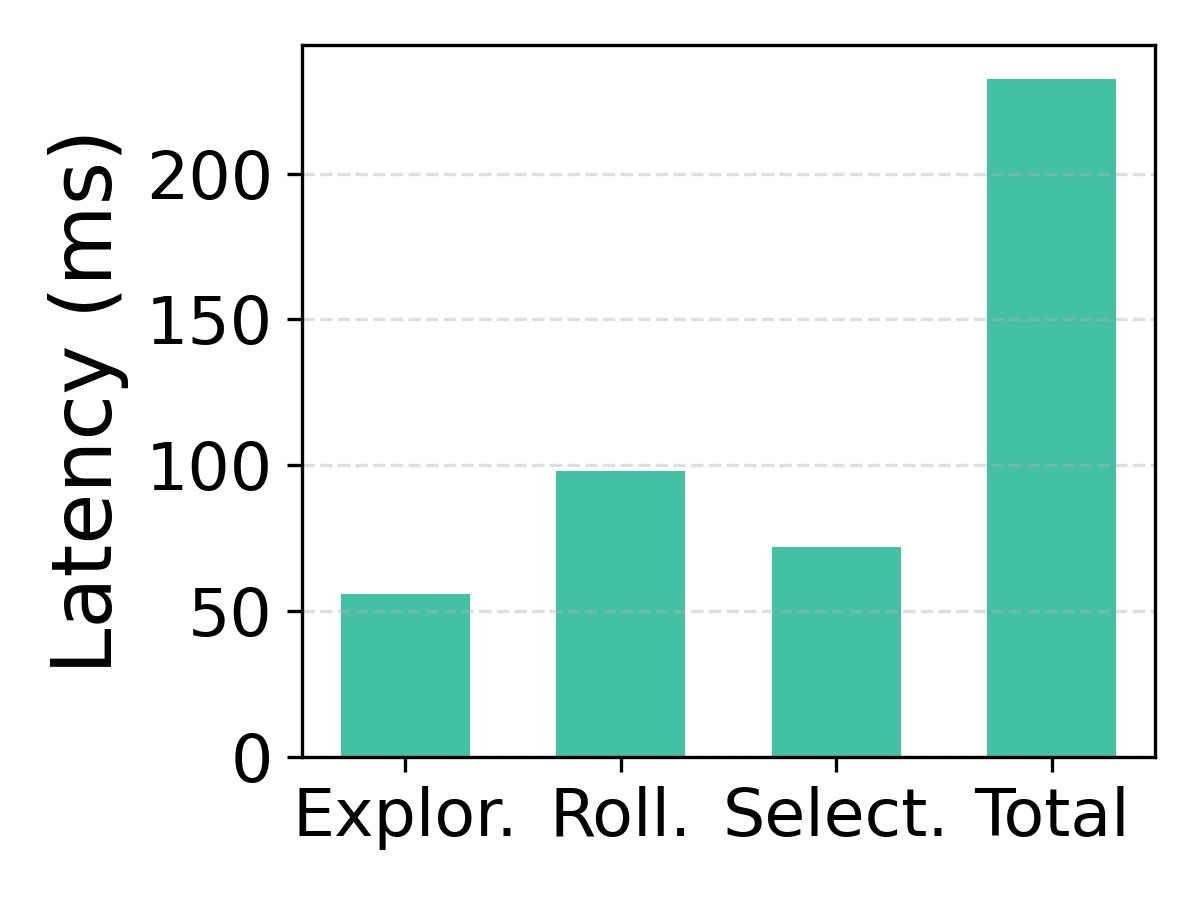}
        \caption{Latency.}
        \label{fig:phone latency overhead}
    \end{subfigure}%
    \begin{subfigure}{0.33\linewidth}
            \setlength{\abovecaptionskip}{0.cm}
        \setlength{\belowcaptionskip}{0.cm}
        \raggedright   
        \includegraphics[width=1\linewidth]{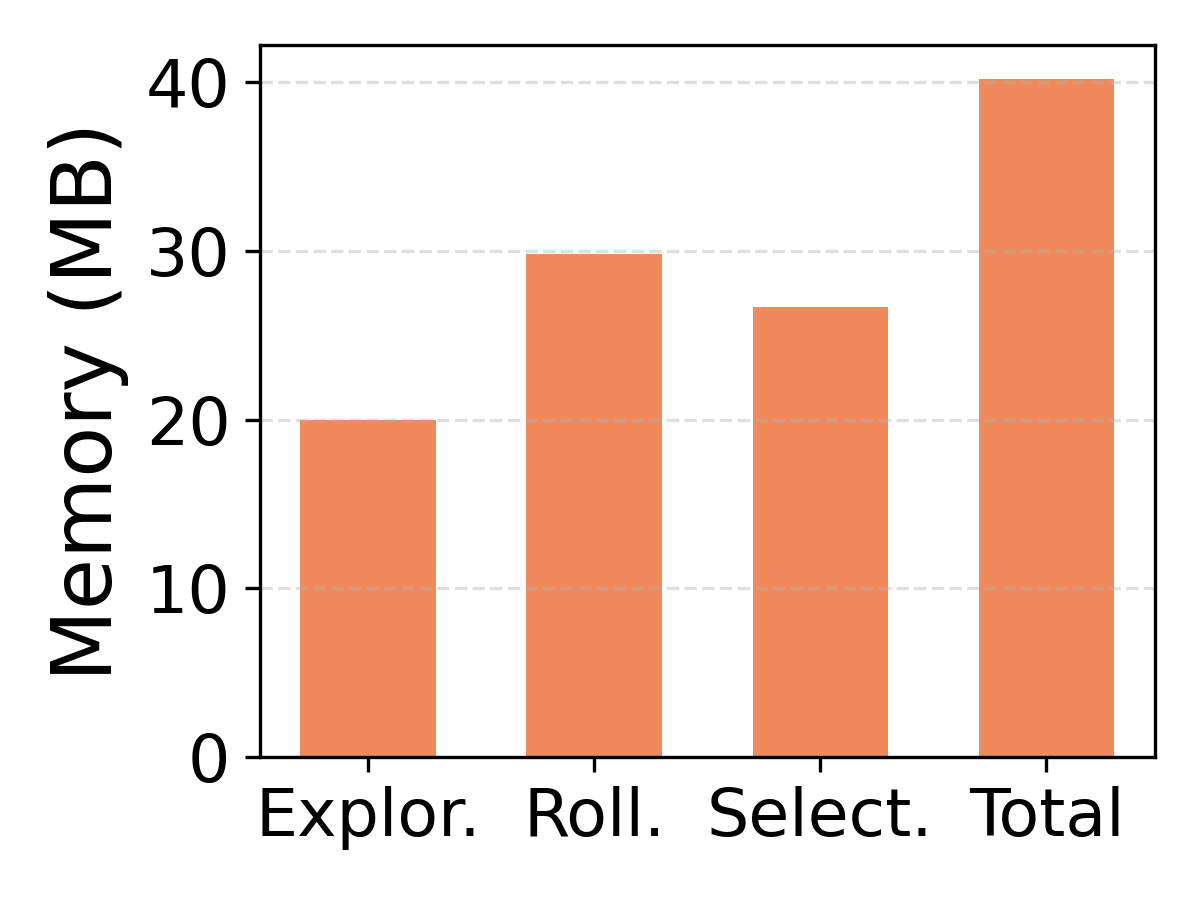}
        \caption{Memory.}
        \label{fig:phone memory overhead}
    \end{subfigure}%
    \begin{subfigure}{0.33\linewidth}
            \setlength{\abovecaptionskip}{0.cm}
        \setlength{\belowcaptionskip}{0.cm}
        \raggedright   
        \includegraphics[width=1\linewidth]{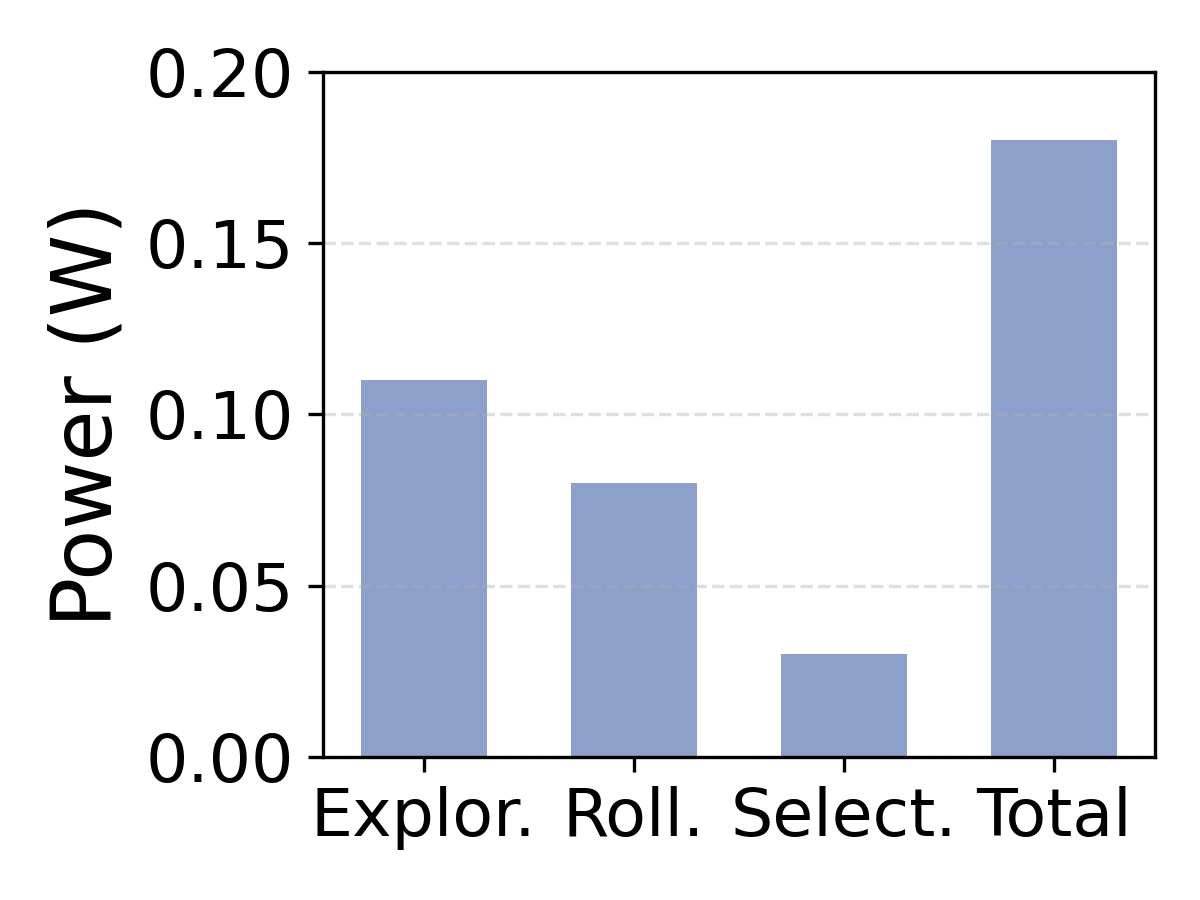}
        \caption{Power.}
        \label{fig:phone power overhead}
    \end{subfigure}%
    
    \caption{Overhead of different components.}
    \label{fig:overhead on phone}
\end{figure}

We further analyze the runtime overhead of individual modules on the smartphone. Fig.~\ref{fig:overhead on phone} shows that the additional control logic introduces only tens of milliseconds of latency per step, which is negligible compared with multi-second VLM inference. Rollback verification incurs the largest latency due to screenshot-based state checking, while exploration and element selection remain lightweight. Memory and power usage also remain stable across modules.

\section{Conclusion and Discussions}
In this paper, we introduce \name, an on-device mobile GUI agent framework that improves task efficiency through task relevance-driven online exploration. Instead of leaving the system idle during expensive VLM reasoning, \name utilizes the reasoning window to probe semantically relevant UI elements and collect additional interface context, while a two-level rollback mechanism ensures that exploration does not disrupt the main execution trajectory. Experimental results show that \name maintains comparable task success while reducing the average number of interaction steps by approximately 23\%, demonstrating that latency-bounded exploration can effectively improve the efficiency of mobile GUI agents. In future work, we will explore more adaptive exploration strategies that dynamically adjust exploration paths based on task demands and UI complexity, and investigate more complicated interplay between UI exploration and model reasoning to further improve the efficiency of on-device GUI agent systems.

\bibliographystyle{ACM-Reference-Format}
\bibliography{reference}

\end{document}